\documentclass[times, review, 10pt]{elsarticle}

\usepackage{xspace}
\usepackage{natbib}
\usepackage{algorithm}
\usepackage{algorithmic}
\usepackage{amssymb}
\usepackage{hyperref}
\usepackage{amsmath}
\usepackage{multirow}
\newcommand{\method}{LoRD\xspace}
\usepackage{subcaption}
\usepackage{algorithm}
\usepackage{algorithmic}
\usepackage{amssymb}
\usepackage{amsmath}
\usepackage{multirow}
\usepackage[title]{appendix}
\def\tsc#1{\csdef{#1}{\textsc{\lowercase{#1}}\xspace}}
\tsc{WGM}
\tsc{QE}
\tsc{EP}
\tsc{PMS}
\tsc{BEC}
\tsc{DE}
\usepackage{xcolor}

\journal{Pattern Recognition}

\begin{document}

\begin{frontmatter}



\title{Low-redundancy Distillation for Continual Learning}


\author[1,2]{Ruiqi Liu}
\author[1,2]{Boyu Diao\corref{cor1}}
\author[1]{Libo Huang}
\author[1,2]{Zijia An}
\author[1,2]{Hangda Liu}
\author[1,2]{Zhulin An}
\author[1,2]{Yongjun Xu}
\affiliation[1]{organization={Institute of Computing Technology},
    addressline={Chinese Academy of Sciences}, 
    city={Beijing},
    postcode={100190}, 
    country={China}}
\affiliation[2]{organization={University of Chinese Academy of Sciences}, 
    city={Beijing},
    postcode={100049}, 
    country={China}}
\cortext[cor1]{Corresponding author}

\begin{abstract}
Continual learning (CL) aims to learn new tasks without erasing previous knowledge. However, current CL methods primarily emphasize improving accuracy while often neglecting training efficiency, which consequently restricts their practical application. Drawing inspiration from the brain's contextual gating mechanism, which selectively filters neural information and continuously updates past memories, we propose Low-redundancy Distillation (\method), a novel CL method that enhances model performance while maintaining training efficiency. This is achieved by eliminating redundancy in three aspects of CL: student model redundancy, teacher model redundancy, and rehearsal sample redundancy. By compressing the learnable parameters of the student model and pruning the teacher model, \method facilitates the retention and optimization of prior knowledge, effectively decoupling task-specific knowledge without manually assigning isolated parameters for each task. Furthermore, we optimize the selection of rehearsal samples and refine rehearsal frequency to improve training efficiency. Through a meticulous design of distillation and rehearsal strategies, \method effectively balances training efficiency and model precision. Extensive experimentation across various benchmark datasets and environments demonstrates \method's superiority, achieving the highest accuracy with the lowest training FLOPs.
\end{abstract}



\begin{keyword}


Continual Learning  \sep  Lifelong Learning \sep Catastrophic Forgetting \sep Knowledge Distillation \sep Experience Replay
\end{keyword}

\end{frontmatter}

\section{Introduction}
\label{sec:intro}

Continual learning is the process where a machine learning model adapts to new data while retaining old knowledge in a dynamic environment\textcolor{cyan}{~\cite{zhou2024class}}. Within CL, Deep Neural Networks face the challenge of \textit{Catastrophic Forgetting}~\cite{mccloskey1989catastrophic}, an issue where the acquisition of new knowledge can lead to a rapid erosion of previously learned knowledge\textcolor{cyan}{~\cite{yang2024continual}}. Although recent methods mainly focus on this forgetting issue, deploying CL on edge devices, such as the NVIDIA Jetson TX2\textcolor{cyan}{~\cite{liu2024resource}}, is crucial for achieving real-time edge intelligence. When deploying CL on resource-constrained devices, learning efficiency becomes equally important. However, prior CL works have often overlooked the inherent efficiency of the algorithms themselves, focusing primarily on improving accuracy.

Various methods have been implemented to mitigate \textit{Catastrophic Forgetting}. Architecture based methods\textcolor{cyan}{~\cite{gurbuz2022nispa}} allocate distinct parameters for each task to decouple task-specific knowledge. Regularization based methods~\cite{yu2020semantic} restrict updates to crucial parameters or distill the entire model. Rehearsal based methods~\cite{buzzega2020dark} prevent forgetting by retaining samples from previous tasks. Although these methods enhance model performance, they introduce considerable computational overhead during training, limiting their practical application—even underperforming the simplest rehearsal-based method ER~\cite{robins1995catastrophic} in real-world applications\textcolor{cyan}{~\cite{prabhu2023computationally}}. A limited number of studies explore training efficiency in CL\textcolor{cyan}{~\cite{liu2024continual}}. Among these, SparCL~\cite{wang2022sparcl} reduces the FLOPs required for model training by implementing dynamic weight and gradient masks, along with selective sampling of crucial data. These methods accelerate the training process through pruning and sparse training but do not achieve joint optimization with precision, resulting in mediocre performance.

Knowledge distillation and sample rehearsal have been proven to be the most effective methods for mitigating \textit{catastrophic forgetting}\textcolor{cyan}{~\cite{li2023memory}}. However, they face three significant challenges: (1) the effectiveness of the distillation methods is compromised because the teacher model cannot encode all useful information efficiently~\cite{boschini2022class}; (2) rehearsal-based methods suffer from an inherent imbalance between samples of previous and current tasks, leading to model updates biased toward the current task  (\textit{i.e.} recency bias~\cite{caccia2021reducing}); (3) both methods introduce substantial computational overhead, resulting in poor performance in real-world applications\textcolor{cyan}{~\cite{ghunaim2023real}}. In contrast, the human brain employs a gating mechanism to filter redundant neural information based on context, allowing for efficient retention and updating of past memories~\cite{kudithipudi2022biological}. Drawing inspiration from this biological mechanism, we aim to address these three challenges by eliminating redundancy in both distillation and rehearsal-based methods. Our goal is to achieve state-of-the-art (SOTA) accuracy while reducing training FLOPs, ultimately improving efficiency compared to ER, which has demonstrated superior performance in real-world applications and serves as a critical benchmark for evaluating efficiency~\cite{prabhu2023computationally}.

\begin{figure} [t!]
     \centering
     \includegraphics[width=0.75\linewidth]{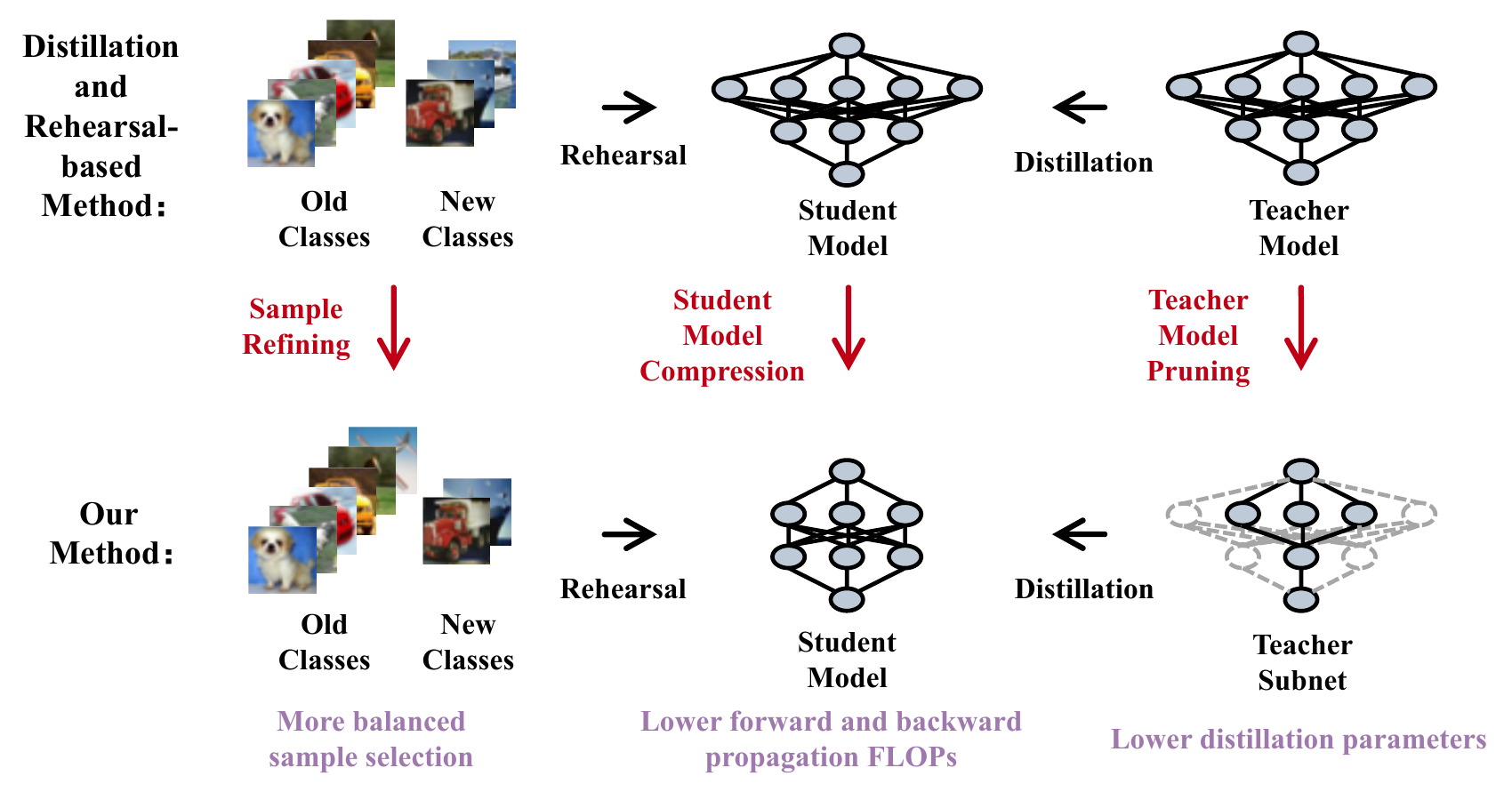}  
     \caption{Overview of \method. We employ Sample Refining to achieve a more balanced sample selection, Student Model Compression to reduce forward and backward propagation FLOPs, and Teacher Model Pruning to minimize distillation parameters. Each of these techniques targets redundancy reduction, leading to improved accuracy and enhanced training efficiency.}
    \label{fig:over}  
\end{figure}
To this end, we propose the Low-redundancy Distillation (\method), a novel CL method that achieves collaborative optimization of accuracy and training efficiency by eliminating redundancy in the learning process. Inspired by the human brain's gating mechanism, which filters out redundant information to efficiently retain and update past memories, \method reduces redundancy in the CL process across three aspects: the student model, the teacher model, and replay samples, as shown in Fig.~\ref{fig:over}. Specifically, we compress the number of learnable parameters in the student model, allocating parameters at each task boundary to ensure its plasticity. Simultaneously, we save the student model from the last task as the teacher model for the current task. The teacher model is pruned to create a teacher subnet, and the subnet with an identical structure in the student model is distilled to maintain its stability. This distillation method not only reduces redundancy in the CL distillation process but also allows non-distilled weights to adapt to new tasks while distilled weights retain knowledge from previous tasks. \method provides a new mechanism for decoupling task-specific knowledge, effectively addressing the issue of recency bias. To help the teacher model efficiently encode and optimize previous knowledge, consistent with DER~\cite{buzzega2020dark}, we retain the logits of previous samples for distillation and utilize two collaborative distillation methods. Furthermore, we optimize the selection of rehearsal samples to further reduce redundancy in the rehearsal process, thereby boosting the model's performance and training efficiency. Aligned with the brain's gating mechanism, \method enables CL models to efficiently filter and preserve crucial knowledge, optimizing it in subsequent learning processes. \method achieves SOTA accuracy with lower training FLOPs than the ER method, facilitating its application in real-world scenarios.

In summary, our contributions are as follows:
\begin{itemize}
\item Inspired by the human brain's gating mechanism, which efficiently filters out redundant information to retain and update past memories, we propose \method, a novel CL method that optimizes model performance while maintaining training efficiency by eliminating redundancy in three aspects of CL: student model redundancy, teacher model redundancy, and rehearsal sample redundancy.
\item We explore the synergy between distillation and rehearsal methods, and as a result, \method effectively decouples task-specific knowledge without manually assigning isolated parameters for each task.
\item Experiments conducted in both cloud and edge environments have demonstrated that \method outperforms SOTA methods, achieving up to a 3.8\% increase in accuracy while reducing training FLOPs by 12\% compared to the ER method.
\end{itemize}
\section{Related Work}
\label{sec:related}

\paragraph{Effective Continual Learning}
Effective continual learning aims to mitigate \textit{catastrophic forgetting} and is typically classified into three main categories: regularization-based methods, architecture-based methods, and rehearsal-based methods.

\textit{Regularization-based Methods}\textcolor{cyan}{~\cite{li2017learning}} limit the drift of crucial network parameters in previous tasks. The distillation methods\textcolor{cyan}{~\cite{huang2024etag}}, which are also considered regularization methods, constrain the new model's output for the original task to be similar to that of the old model by designating the previous model snapshot as the teacher. However, the teacher model may encounter forgetting problems, a challenge effectively tackled by \method through the synergistic utilization of two distinct types of distillation losses.

\textit{Architecture-based Methods}\textcolor{cyan}{~\cite{sun2023exemplar}} allocate distinct parameter sets for each task. Certain architecture-based methods~\cite{gurbuz2022nispa} focus on freezing or pruning subnets to decouple knowledge. \method enhances the model's stability through distillation and sample replay, avoiding the manual selection and freezing of crucial weights, thereby promoting the model's forward transfer capability~\cite{lopez2017gradient}. Moreover, \method improves the model's plasticity by assigning parameters to the student model instead of persistently pruning it.

\textit{Rehearsal-based Methods}\textcolor{cyan}{~\cite{wang2025continual}} employ a replay buffer to store and periodically revisit data from previous tasks, preserving knowledge and mitigating forgetting.
ER~\cite{robins1995catastrophic} enhances learning by interleaving current task samples with those from previous tasks. Building upon this, DER~\cite{buzzega2020dark} further augments the learning process by storing past model logits and utilizing these logits for distillation. Although rehearsal-based methods are acknowledged as the current state-of-the-art~\cite{buzzega2020dark}, their effectiveness can be severely impacted if the buffer size is small. Additionally, prior studies~\cite{prabhu2023computationally} have found that the training FLOPs of CL methods are crucial for real-world applications because long training times before deployment are impractical. Consequently, the ER method often achieves the best performance in practical settings due to its low training FLOPs. \method improves training efficiency by eliminating redundancy, achieving SOTA performance with lower training FLOPs than ER.
\begin{figure*} [!t]
     \centering
     \includegraphics[width=0.95\linewidth]{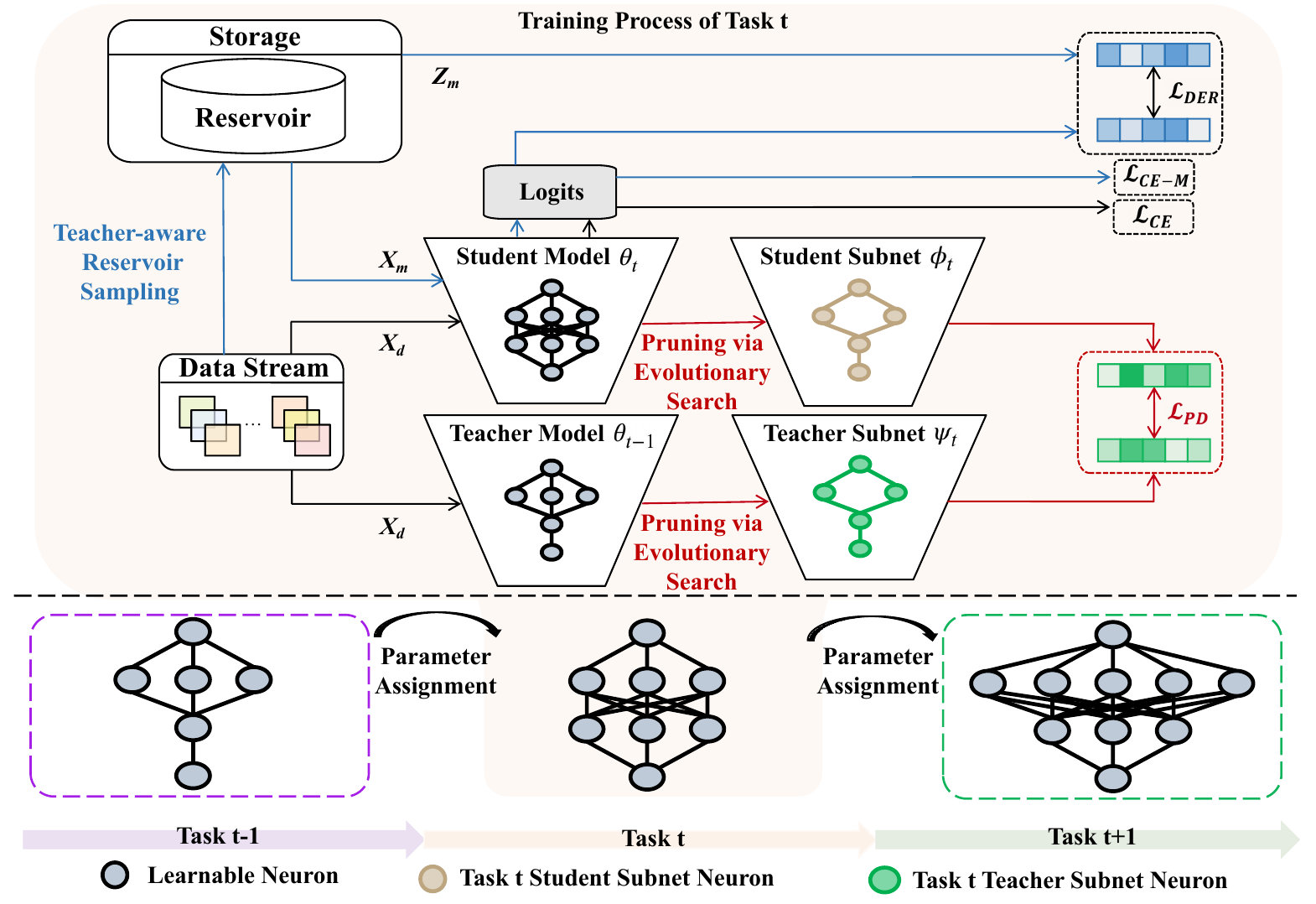}  
     \caption{The main framework of \method.
        \textbf{Above the dashed line:} During the training of each task, we prune the teacher model into a teacher subnet and distill the student subnet that has the same structure as the teacher subnet. Additionally, we propose Teacher-aware Reservoir Sampling to optimize the selection of replay samples.
        \textbf{Below the dashed line:} We compress the number of learnable parameters in the student model to reduce redundancy. At each task boundary, we assign these learnable parameters to the student model to ensure its plasticity.}
    \label{fig:method}  
\end{figure*}
\paragraph{Efficient Continual Learning}
The aim of efficient continual learning\textcolor{cyan}{~\cite{harun2023siesta}} is to enhance the efficiency of CL by optimizing training speed, memory usage, and computational resource utilization. SparCL~\cite{wang2022sparcl} stands out among various methods by reducing the model's training FLOPs through the utilization of dynamic weight and gradient masks, along with selective sampling of crucial data. In contrast to conventional efficient CL methods that primarily focus on optimizing training efficiency, \method aims to maximize model performance while maintaining training efficiency by eliminating redundancy in the learning process. Furthermore, our method can be combined with efficient CL methods, as demonstrated in Sec.~\ref{sec:exp:sparse}.

\section{Method}
\label{sec:method}
We describe a novel CL method, \method (as shown in Fig.~\ref{fig:method}), in this section. \method aims to optimize model performance while maintaining training efficiency by eliminating redundancy in the learning process. Before going into in-depth, we first clarify the problem setting.
\subsection{Problem Setting}
\label{sec:method:overview}
The CL problem involves sequentially learning $T$ tasks from a dataset, where each task $t$ has a training set $\mathcal{D}^t=\{(x_i^t,y_i^t)\}_{i=1}^{N_t}$ with $N_t$ i.i.d. samples represented by pairs $(x_i^t,y_i^t)$.
The objective of CL is to perform sequential training while preserving the performance of previous tasks. This objective can be achieved by learning a working model $f_\theta(\cdot)$ that minimizes classification loss across all tasks $t=\{1,..., T\}$~\cite{buzzega2020dark}:
\begin{equation}
    \mathcal{L}_\text{CE} =  \mathbb{E}_{(x^t,y^t)\sim \mathcal{D}^t} \ell(\sigma(f_\theta (x)), y),
    \label{eq:ce-loss}
\end{equation}
where $\ell(\cdot, \cdot)$ represents cross-entropy loss, $\sigma(\cdot)$ is the softmax function. Due to the unavailability of historical task data, rehearsal-based methods incorporate a replay buffer $\mathcal{M}={(x_i,y_i)}_{i=1}^{\mathcal{B}}$. This buffer, with a constant size $\mathcal{B}$, facilitates adapting to new tasks and retaining knowledge from previous tasks.
During the training process, we replay samples from the buffer based on ER~\cite{robins1995catastrophic}. The classification loss on memory can be formulated as:
\begin{equation}
    \mathcal{L}_\text{CE-M} =  \mathbb{E}_{(x,y)\sim \mathcal{M}} \ell(\sigma(f_\theta (x)), y).
    \label{eq:ce-M-loss}
\end{equation}
Furthermore, the distillation methods utilize logit matching~\cite{hinton2015distilling} to efficiently reduce changes in the crucial parameters of the model. Among these methods, DER~\cite{buzzega2020dark} stores the output logits $z \triangleq f_{\theta^t}(x)$ of previous models and utilizing them for distillation, where $t$ indicates the time of memory insertion:
\begin{equation}
    \mathcal{L}_\text{DER} =  \mathbb{E}_{(x,z)\sim \mathcal{M}} \| f_\theta (x), z \|_2^2,
    \label{eq:der-loss}
\end{equation}
and DER++~\cite{buzzega2020dark} further replays both logits and ground-truth labels:
\begin{equation}
    \mathcal{L}_\text{DER++} =  \beta_1\mathcal{L}_\text{CE-M}+\beta_2\mathcal{L}_\text{DER},
    \label{eq:derpp-loss}
\end{equation}
where $\beta_1$ and $\beta_2$ are weighting factors.
\subsection{Low-redundancy Distillation}
The human brain possesses a gating mechanism that filters out redundant neural information based on context and continuously updates past memories. Therefore, we propose pruning through evolutionary search and compression via cosine annealing in the distillation process to reduce redundancy. In \method, we use the working model from the last task as the teacher model $f_{\theta_{t-1}}(\cdot)$ for the current task, while the current model serves as the student model $f_{\theta_{t}}(\cdot)$. By filtering out redundancies, \method enables the teacher model to consistently update and preserve essential knowledge.
\subsubsection{Student Model Compression}
\label{sec:method:masking}
We reduce the redundancy of the student model by compressing its learnable parameters. Instead of directly updating the entire student model, we incrementally allocate the learnable parameters. We segment the filters in each layer into $G$ groups~\cite{cai2020once}. Subsequently, at the boundary of the $t$-th task, we select $s_t$ groups of filters to assign as learnable filters for the student model. The allocation of parameters is crucial, as we need to leave a room for future tasks while fully learning the current task. For initial task learning, \method employs knowledge distillation to concentrate previous task knowledge within the distilled student subnet, while enabling non-distilled parameters to rapidly adapt to new tasks. However, with increasing task numbers, the distilled student subnet demonstrates gradual forgetting of prior knowledge, while replay methods tend to utilize more non-distilled parameters for old knowledge consolidation, consequently impairing model plasticity. To address this, \method incorporates a cosine annealing algorithm for parameter allocation, thereby assigning supplementary parameters to subsequent tasks to maintain plasticity. This compression method allows the student model to ensure stability through replay and distillation mechanisms while gaining additional parameters to enhance plasticity, thus effectively addressing catastrophic forgetting with incremental tasks while improving training efficiency through optimized parameter utilization. For the $t$-th task boundary, the $s_t$ is determined as follows:
\begin{equation}
    s_t=\frac{1}{2} r\left(1+\cos \left(\frac{t \times \pi}{N}\right)\right),
    \label{eq:cosine_annealing}
\end{equation}
where $0 \leq t \leq N-1$ and $N$ is the expected number of tasks for CL. To fully leverage the capacity of the student model, all parameters should be learnable upon reaching the $N$-th task boundary. The value of $r$ is thus determined as follows:
\begin{equation}
    r=\frac{2 G}{\sum_{t=0}^{N-1}\left(1+\cos \left(\frac{t \times \pi}{N}\right)\right)},
    \label{eq:r_definition}
\end{equation}
and the number of learnable filter groups $g_t$ for the $t$-th task can be calculated as follows:
\begin{equation}
    g_t = \sum_{i=0}^{t-1}s_i = \frac{G \sum_{i=0}^{t-1} \left(1+\cos \left(\frac{i \times \pi}{N}\right)\right)}{\sum_{t=0}^{N-1}\left(1+\cos \left(\frac{t \times \pi}{N}\right)\right)}.
    \label{eq:learnable_parameters}
\end{equation}
When the number of tasks $N$ is unknown (i.e., when $N$ approaches infinity), $g_t$ becomes infinitesimally small. Therefore, we constrain $s_t$ to have a minimum value of 1. Our parameter allocation method controls the learnable parameters in the student model through structured pruning, which achieves model compression by systematically removing entire structural components rather than applying parameter masking. Through this parameter allocation process, \method dynamically generates authentic student subnets, thereby ensuring reduced computational complexity during both forward and backward propagation.

\subsubsection{Teacher Model Pruning}
\label{sec:method:pruning}
We utilize the student model from the previous task as the teacher model for the current task and prune it to reduce redundancy. In contrast to conventional pruning methods~\cite{wang2022sparcl} that consider weight importance, we design pruning strategies based on three perspectives: (1) We prune filters to obtain a teacher subnet to enhance distillation efficiency. (2) The pruned teacher subnet should have similar logits outputs to the teacher network, and (3) the redundancy of the teacher subnet should be reduced, meaning the number of parameters should be small. We formalize the problem as follows:
\begin{equation}
    \underset{\mathcal{\psi}_t}{\operatorname{argmin}} \quad \textit{exp}(\frac{|\psi_t|}{|\theta_{t-1}|})\| f_{\theta_{t-1}}(x) - f_{\psi_{t}}(x) \|_2^2,
    \label{eq:prune}
\end{equation}
where $f_{\psi_{t}}(\cdot)$ is the pruned teacher subnet and $|\psi_{t}|$ represents the number of its parameters. $\exp(\frac{|\psi_t|}{|\theta_{t-1}|})$ is utilized to control the size of the parameters. Consistent with~\cite{cai2020once}, we conduct an evolutionary search~\cite{real2019regularized} to get a teacher subnet that minimizes Eq.~\ref{eq:prune}. We define the width $g_t$ of the current student model as the overall search space for the evolutionary search. To construct the initial population of candidate subnets, we randomly select a width value no greater than $g_t$ for each layer of the network architecture. Finally, we employ evolutionary search based on Eq.~\ref{eq:prune} to identify the optimal teacher subnet. In the evolutionary search for the teacher subnet, instead of using masks to record which neurons are activated, we record the width of each layer and obtain the teacher subnet by pruning the less important filters according to the predefined width. The search process completes within a minute.

During the distillation stage, the teacher subnet and the student model are not directly utilized. Instead, we prune a student subnet with a structure identical to the teacher subnet from the student model and perform distillation between these two subnets. We compute the following loss term:
\begin{equation}
    \mathcal{L}_\text{PD} =  \mathbb{E}_{(x,y)\sim \mathcal{D}} \| f_{\phi_{t}}(x) - f_{\psi_{t}}(x) \|_2^2 ,
    \label{eq:ED_loss}
\end{equation}
where $f_{\phi_{t}}(\cdot)$ is the pruned student subnet. By employing Eq.~\ref{eq:ED_loss}, \method allows non-distilled weights to adapt to new tasks, while distilled weights retain knowledge from previous tasks. This method decouples task-specific knowledge and addresses the issue of recency bias. Furthermore, We employ two types of distillation losses, Eq.~\ref{eq:der-loss} and Eq.~\ref{eq:ED_loss}. Relying solely on Eq.~\ref{eq:der-loss} would prevent the effective update of prior knowledge, leading to incomplete encoding of useful information. Conversely, relying solely on Eq.~\ref{eq:ED_loss} would subject the teacher model to forgetting. Through the combined application of both distillation functions, the teacher network can efficiently encode useful knowledge and improve distillation performance.

\begin{algorithm*}[t!]
	\caption{TRS Algorithm}
	\label{alg:reservoir}
	\begin{algorithmic}
		\STATE {\bfseries Input:} Memory Buffer $\mathcal{M}$, Memory Budget $\mathcal{B}$, Number of seen examples $N$, Teacher subnet parameter quantity $|\psi_{t}|$, Student model parameter quantity $|\theta_{t}|$, Selected example $(x, y)$.
		\IF{$\mathcal{B} > N$}  
		\STATE $\mathcal{M}[N] \gets (x,y)$ 
		\hfill $\rhd$ Memory is not full  \ELSE 
		\STATE $k = \text{\textit{randomInteger}}~(\min=0,\max=N)$
        \STATE $i = \text{\textit{random}}~(\min=0,\max=1)$  
		\IF{ $i < \textit{exp}(-\alpha\frac{|\psi_{t}|}{|\theta_{t}|})$ }
        \IF{ $k < \mathcal{B}$ }
		\STATE $\mathcal{M}[k] \gets (x,y)$
		\hfill $\rhd$ Select a sample to remove
        \ENDIF
		\ENDIF
		\ENDIF
		\STATE \textbf{return} $\theta$
	\end{algorithmic}
\end{algorithm*}
\subsection{Teacher-aware Reservoir Sampling}
Here, we provide the sampling algorithm for the TRS.
This algorithm effectively manages the memory buffer based on the knowledge retention capabilities of the teacher subnet, ensuring that valuable experiences are retained. 
\label{sec:method:Replay}
Some prior CL works improve reservoir sampling by selecting informative examples to construct the replay buffer~\cite{yoon2022online}. However, they do not consider computing resource utilization~\cite{aljundi2019gradient}, resulting in excessive computational costs.
To enhance replay efficiency while minimizing additional computational burden, we propose Teacher-aware Reservoir Sampling (TRS) method. Due to the use of two types of distillation losses, the purpose of replaying samples has shifted from preventing model forgetting to helping teacher networks better encode knowledge from old tasks. Based on this, we reduce the likelihood of storing the current sample based on the knowledge retention capabilities of the teacher subnet. The probability of storing the $k$-th ($k >\mathcal{B}$) sample in the buffer is denoted as $p(x_k)$ and the calculation of $p(x_k)$ can be performed as follows:
\begin{equation}
    p(x_k) = \textit{exp}(-\alpha\frac{|\psi_{t}|}{|\theta_{t}|})\frac{\mathcal{B}}{k},
    \label{eq:pk}
\end{equation}
where $ \alpha $ is a weighting factor and we assess the knowledge retention capabilities of the teacher subnet based on their size. The sampling process for the TRS algorithm is detailed in Algorithm~\ref{alg:reservoir}, where both indices i and k are sampled from a uniform distribution\footnote{We provide more details about TRS in Appendix~\ref{sec:app:TRS}.}. Furthermore, we delve deeper into examining the correlation between rehearsal frequency (RF) and distillation. Within the DER method, each batch comprises half the samples from the new task and half the samples from the buffer. Notably, the inclusion of the teacher model has resulted in a decreased requirement for replay samples, enabling \method to adjust RF based on the desired trade-off between accuracy and efficiency. We conduct a comprehensive exploration study on RF, detailed in Sec.~\ref{sec:exp:RF}, to further underscore our method's effectiveness.

\begin{algorithm}[!t]
	\caption{The training algorithm of the \method}
	\label{alg:alg_Representative}
	\begin{algorithmic}[1]
		\REQUIRE Data stream $\mathcal{D}$, Buffer $\mathcal{M}$, Model weight $\theta$
        \FOR{$t = 0,...,T$}
        \STATE Assign learnable parameters by Eq.~\ref{eq:learnable_parameters} 
    		\WHILE{Training one task}
            \STATE $\left(X_{d}, Y_{d}\right) \leftarrow \textit{sample}(\mathcal{D})$
            \STATE Calculate $\mathcal{L}_\text{CE}$ and $\mathcal{L}_\text{PD}$ by Eq.~\ref{eq:ce-loss} and Eq.~\ref{eq:ED_loss}
            \STATE $\left(X_{m}, Y_{m}, Z_{m}\right) \leftarrow \textit{sample}(\mathcal{M})$
    		\STATE Calculate $\mathcal{L}_\text{DER++}$ and $\mathcal{L}$ by Eq.~\ref{eq:derpp-loss} and Eq.~\ref{eq:loss}
    		\STATE $\theta \leftarrow \theta-\eta \nabla_{\theta} \mathcal{L}$
    		\STATE $\mathcal{M} \leftarrow$ TRS $\left(\mathcal{M},\left(X_{d}, Y_{d}\right)\right)$
    		\ENDWHILE
        \STATE Prune the teacher model for the next task by Eq.~\ref{eq:prune} \textbf{}
        \ENDFOR
		\STATE \textbf{return} $\theta$
	\end{algorithmic}
\end{algorithm}

\subsection{The Overall Loss}
\label{sec:method:Loss}
In summary, the total training objective for \method can be expressed by:
\begin{align}
    \label{eq:loss}
        \mathcal{L} = \mathcal{L}_\text{CE} + \lambda \mathcal{L}_\text{PD} + \mathcal{L}_\text{DER++},
\end{align}
where $\lambda$ is a weighting factor and the cross-entropy loss $\mathcal{L}_\text{CE}$ focuses on learning the current data; 
$\mathcal{L}_\text{PD}$ assists the model in preserving and updating essential knowledge.; 
$\mathcal{L}_\text{DER++}$ helps teachers model better encoding knowledge.

We provide a detailed description of the overall training procedure of \method in Algorithm~\ref{alg:alg_Representative}.
Upon the arrival of new data, our method involves the following steps:
Firstly, we compute the overall objective $\mathcal{L}$. Subsequently, we update the weights in the compressed student model through gradient descent. Finally, we utilize TRS to update the buffer.

\section{Experiments}
\label{sec:exp}
\subsection{Dataset and Hardware}
\label{sec:exp:dataset}
We conduct experiments on commonly used public datasets, including 
Split CIFAR-10 (S-CIFAR-10)~\cite{buzzega2020dark}, Split CIFAR-100 (S-CIFAR-100)~\cite{chaudhry2019efficient}, and Split Tiny ImageNet (S-Tiny-ImageNet)~\cite{chaudhry2019continual}.
The Split CIFAR-10 dataset comprises 5 tasks, each containing 2 classes, while the Split CIFAR-100 dataset consists of 20 tasks, each containing 5 classes. The Split Tiny ImageNet dataset is created by dividing the original Tiny ImageNet dataset~\cite{le2015tiny}, which includes 200 classes, into 10 tasks, with each task consisting of 20 classes.
We conduct comprehensive experiments utilizing the NVIDIA GTX 2080Ti GPU paired with the Intel Xeon Gold 5217 CPU, as well as the NVIDIA Jetson TX2, boasting 8 GB LPDDR4 Main Memory and 32 GB of eMMC Flash memory.

\subsection{Evaluation Metrics}
We use the following two metrics to measure the performance of various methods:
\begin{equation}
  ACC_t = \frac{1}{t}\sum_{\tau=1}^{t}R_{t,\tau}
  \label{eq:average accuracy}
\end{equation}
\begin{equation}
    FF_t = \frac{1}{t-1} \sum_{j=1}^{t-1} \max_{i \in \{1, \ldots, t-1\}} (R_{i, j} - R_{t, j}),
    \label{eq:forgetting}
\end{equation}
We denote the classification accuracy on the $\tau$-th task after training on the t-th task as $R_{t,\tau}$.
A higher ACC value indicates superior model performance, while a lower FF value signifies enhanced anti-forgetting efficacy of the model.
We employ the Split CIFAR-10, Split CIFAR-100, and Split Tiny ImageNet datasets to validate the effectiveness of our method in both Task Incremental Learning (Task-IL) and Class Incremental Learning (Class-IL)~\cite{huang2024etag} settings.
In the simplest testing setup, it is assumed that the identity of each incoming test instance is known, which is referred to as Task-IL. If, during CL inference, the subset of classes for each sample is not recognized, the scenario transitions to a more complex Class-IL setting. This research primarily focuses on the more intricate Class-IL setting, while the performance of Task-IL is used solely for comparative analysis.

\subsection{Baselines.}
We compare \method with several representative baseline methods, including three regularization-based methods: oEWC~\cite{schwarz2018progress}, SI~\cite{zenke2017continual} and LwF~\cite{li2017learning}, as well as eleven rehearsal-based methods: ER~\cite{robins1995catastrophic}, GEM~\cite{lopez2017gradient}, A-GEM~\cite{chaudhry2019efficient}, iCaRL~\cite{rebuffi2017icarl}, GSS~\cite{aljundi2019gradient}, HAL~\cite{chaudhry2021using}, DER~\cite{buzzega2020dark}, DER++~\cite{buzzega2020dark}, $Co^2L$~\cite{cha2021co2l},
ER-ACE~\cite{caccianew} and SCoMMER~\cite{sarfraz2023sparse}.
It's important to highlight that many rehearsal-based methods also integrate regularization.
In our evaluation, we incorporate two non-continual learning baselines: SGD (lower bound) and JOINT (upper bound).

\subsection{Implementation Details}
For the CIFAR and Tiny ImageNet datasets, we adopt a standard ResNet18~\cite{he2016deep} architecture without pretraining as the baseline, following the method taken in~\cite{rebuffi2017icarl}.
We expand the Mammoth CL repository in PyTorch~\cite{buzzega2020dark}, and we employ random crops and horizontal flips as data augmentation techniques for both examples from the current task and the replay buffer.
All models are trained using the Stochastic Gradient Descent (SGD) optimizer and the batch size is fixed at 32. The
details of other hyperparameters can be found in Appendix~\ref{sec:app:Selection} and Appendix~\ref{sec:app:Hyperparameter}. In the Split Tiny ImageNet setting, the models are trained for 100 epochs. As for the Split CIFAR-10 and Split CIFAR-100 settings, the models are trained for 50 epochs in each phase.
For rehearsal-based methods, each batch comprises a combination of half the samples from the new task and half the samples from the buffer. When the RF is set to 1/2, the calculation of the loss on the buffer for backpropagation will occur only once every two batches. To ensure robustness, all experiments are repeated 10 times with different initializations, and the results are averaged across the runs.
\begin{table*}[t!]
    \centering
    \small
    \renewcommand\arraystretch{1.2} \setlength{\tabcolsep}{4mm}
    \caption{Classification results for S-CIFAR-10 dataset. The highest results are marked in bold.}
    \begin{tabular}{ccccccccc}
        \hline
        \multirow{2}{*}{{\bf Method}} & \multicolumn{2}{c}{{\bf Class-IL}} & \multicolumn{2}{c}{{\bf Task-IL}} \\
        \cline{2-5}
        & 200 & 500 & 200 & 500 \\
        \hline
        JOINT & \multicolumn{2}{c}{92.20\scriptsize{$\pm$0.15}} & \multicolumn{2}{c}{98.31\scriptsize{$\pm$0.12}} \\
        \hline
        SGD & \multicolumn{2}{c}{19.62\scriptsize{$\pm$0.05}} & \multicolumn{2}{c}{61.02\scriptsize{$\pm$3.33}} \\
        \hline
        GEM~\cite{lopez2017gradient} & 25.54\scriptsize{$\pm$0.76} & 26.20\scriptsize{$\pm$1.26} & 90.44\scriptsize{$\pm$0.94} & 92.16\scriptsize{$\pm$0.69} \\
        \hline
        iCaRL~\cite{rebuffi2017icarl} & 49.02\scriptsize{$\pm$3.20} & 47.55\scriptsize{$\pm$3.95} & 88.99\scriptsize{$\pm$2.13} & 88.22\scriptsize{$\pm$2.62} \\
        \hline
        ER~\cite{robins1995catastrophic} & 44.79\scriptsize{$\pm$1.86} & 57.74\scriptsize{$\pm$0.27}  & 91.19\scriptsize{$\pm$0.94} &  93.61\scriptsize{$\pm$0.27} \\
        \hline
        A-GEM~\cite{chaudhry2019efficient} & 20.04\scriptsize{$\pm$0.34} & 22.67\scriptsize{$\pm$0.57} & 83.88\scriptsize{$\pm$1.49} & 89.48\scriptsize{$\pm$1.45} \\
        \hline
        GSS~\cite{aljundi2019gradient} & 39.07\scriptsize{$\pm$5.59} & 49.73\scriptsize{$\pm$4.78} & 88.80\scriptsize{$\pm$2.89} & 91.02\scriptsize{$\pm$1.57} \\
        \hline
        DER~\cite{buzzega2020dark} & 61.93\scriptsize{$\pm$1.79} & 70.51\scriptsize{$\pm$1.67} & 91.40\scriptsize{$\pm$0.92} & 93.40\scriptsize{$\pm$0.39} \\
        \hline
        DER++~\cite{buzzega2020dark} & 64.88\scriptsize{$\pm$1.17} & 72.70\scriptsize{$\pm$1.36} & 91.92\scriptsize{$\pm$0.60} & 93.88\scriptsize{$\pm$0.50} \\
        \hline
        HAL~\cite{chaudhry2021using} & 32.36\scriptsize{$\pm$2.70} & 41.79\scriptsize{$\pm$4.46} & 82.51\scriptsize{$\pm$3.20} & 84.54\scriptsize{$\pm$2.36} \\
        \hline
        $Co^2L$~\cite{cha2021co2l} & 65.57\scriptsize{$\pm$1.37} & 74.26\scriptsize{$\pm$0.77} & 93.43\scriptsize{$\pm$0.78} & {95.9\scriptsize{$\pm$0.26}} \\
        \hline
        ER-ACE~\cite{caccianew} & 63.22\scriptsize{$\pm$1.44} & 71.85\scriptsize{$\pm$0.52} & 92.28\scriptsize{$\pm$0.34} & 94.22\scriptsize{$\pm$0.18} \\
        \hline
        SCoMMER~\cite{sarfraz2023sparse} & 69.19\scriptsize{$\pm$0.61} & {74.97\scriptsize{$\pm$1.05}} & 93.20\scriptsize{$\pm$0.10} & 94.36\scriptsize{$\pm$0.06} \\
        \hline
        {\bf \method(Ours)} & {\bf 70.16\scriptsize{$\pm$1.23}} & \bf  75.68\scriptsize{$\pm$0.53} & { 93.94\scriptsize{$\pm$0.30}} &   94.97\scriptsize{$\pm$0.29} \\
        \hline
    \end{tabular}
    \label{tab:s-cifar10}
\end{table*}
\begin{table*}[t!]
    \centering
    \small
    \renewcommand\arraystretch{1.2} \setlength{\tabcolsep}{4mm}
    \caption{Classification results for S-Tiny-ImageNet dataset. The highest results are marked in bold.}
    \begin{tabular}{ccccccccc}
        \hline
        \multirow{2}{*}{{\bf Method}} & \multicolumn{2}{c}{{\bf Class-IL}} & \multicolumn{2}{c}{{\bf Task-IL}} \\
        \cline{2-5}
        & 200 & 500 & 200 & 500 \\
        \hline
        JOINT & \multicolumn{2}{c}{59.99\scriptsize{$\pm$0.19}} & \multicolumn{2}{c}{82.04\scriptsize{$\pm$0.10}} \\
        \hline
        SGD & \multicolumn{2}{c}{7.92\scriptsize{$\pm$0.26}} & \multicolumn{2}{c}{18.31\scriptsize{$\pm$0.68}} \\
        \hline
        GEM~\cite{lopez2017gradient} & 12.56\scriptsize{$\pm$0.41} & 14.92\scriptsize{$\pm$1.06} & 39.28\scriptsize{$\pm$0.77} & 48.98\scriptsize{$\pm$0.89} \\
        \hline
        iCaRL~\cite{rebuffi2017icarl} & 7.53\scriptsize{$\pm$0.79} & 9.38\scriptsize{$\pm$1.53} & 28.19\scriptsize{$\pm$1.47} & 31.55\scriptsize{$\pm$3.27} \\
        \hline
        ER~\cite{robins1995catastrophic} & 8.49\scriptsize{$\pm$0.16} & 9.99\scriptsize{$\pm$0.29} & 38.17\scriptsize{$\pm$2.00} &  48.64\scriptsize{$\pm$0.46} \\
        \hline
        A-GEM~\cite{chaudhry2019efficient} & 8.07\scriptsize{$\pm$0.08} & 8.06\scriptsize{$\pm$0.04} & 22.77\scriptsize{$\pm$0.03} & 25.33\scriptsize{$\pm$0.49} \\
        \hline
        GSS~\cite{aljundi2019gradient} & 14.78\scriptsize{$\pm$0.87} & 18.23\scriptsize{$\pm$1.35} & 38.22\scriptsize{$\pm$0.98} & 50.47\scriptsize{$\pm$0.72} \\
        \hline
        DER~\cite{buzzega2020dark} & 11.87\scriptsize{$\pm$0.78} & 17.75\scriptsize{$\pm$1.14} & 40.22\scriptsize{$\pm$0.67} & 51.78\scriptsize{$\pm$0.88} \\
        \hline
        DER++~\cite{buzzega2020dark} & 10.96\scriptsize{$\pm$1.17} & 19.38\scriptsize{$\pm$1.41} & 40.87\scriptsize{$\pm$1.16} & 51.91\scriptsize{$\pm$0.68} \\
        \hline
        HAL~\cite{chaudhry2021using} & 13.07\scriptsize{$\pm$0.57} & 16.95\scriptsize{$\pm$0.89} & 39.95\scriptsize{$\pm$1.12} & 51.36\scriptsize{$\pm$0.58} \\
        \hline
        $Co^2L$~\cite{cha2021co2l} & 13.88\scriptsize{$\pm$0.4} & 20.12\scriptsize{$\pm$0.42} & 42.37\scriptsize{$\pm$0.74} & 53.04\scriptsize{$\pm$0.69} \\
        \hline
        ER-ACE~\cite{caccianew} & 13.82\scriptsize{$\pm$0.03} & 21.16\scriptsize{$\pm$0.15} & 45.09\scriptsize{$\pm$0.25} & 54.39\scriptsize{$\pm$0.35} \\
        \hline
        SCoMMER~\cite{sarfraz2023sparse} & 9.50\scriptsize{$\pm$0.86} & 17.85\scriptsize{$\pm$0.35} & 35.95\scriptsize{$\pm$3.02} & 52.85\scriptsize{$\pm$0.67} \\
        \hline
        {\bf \method(Ours)} & {\bf 17.68\scriptsize{$\pm$1.06}} & {\bf  24.01\scriptsize{$\pm$0.36}} & {\bf 47.68\scriptsize{$\pm$1.23}} & {\bf  50.08\scriptsize{$\pm$0.74}} \\
        \hline
    \end{tabular}
    \label{tab:s-tiny-imagenet}
\end{table*}
\subsection{Experimental Results}
\label{sec:exp:Results}

\paragraph{Buffer-based Settings}
Tab.~\ref{tab:s-cifar10} and Tab.~\ref{tab:s-tiny-imagenet} present the ACC of all tasks after training. The results demonstrate that \method achieves superior performance across different experimental settings. In Class-IL, \method outperforms the second-best method by up to 3.8\%. Conventional rehearsal-based methods predominantly depend on replay samples to retain knowledge from previous tasks. Conversely, our method utilizes low-redundancy distillation to decouple task-specific knowledge, allowing \method to achieve superior performance.

\paragraph{Buffer-free Settings.}
\label{sec:exp:Results}
\begin{figure}[t]
    \centering
    \begin{subfigure}[b]{0.4\textwidth}
      \includegraphics[width=\textwidth]{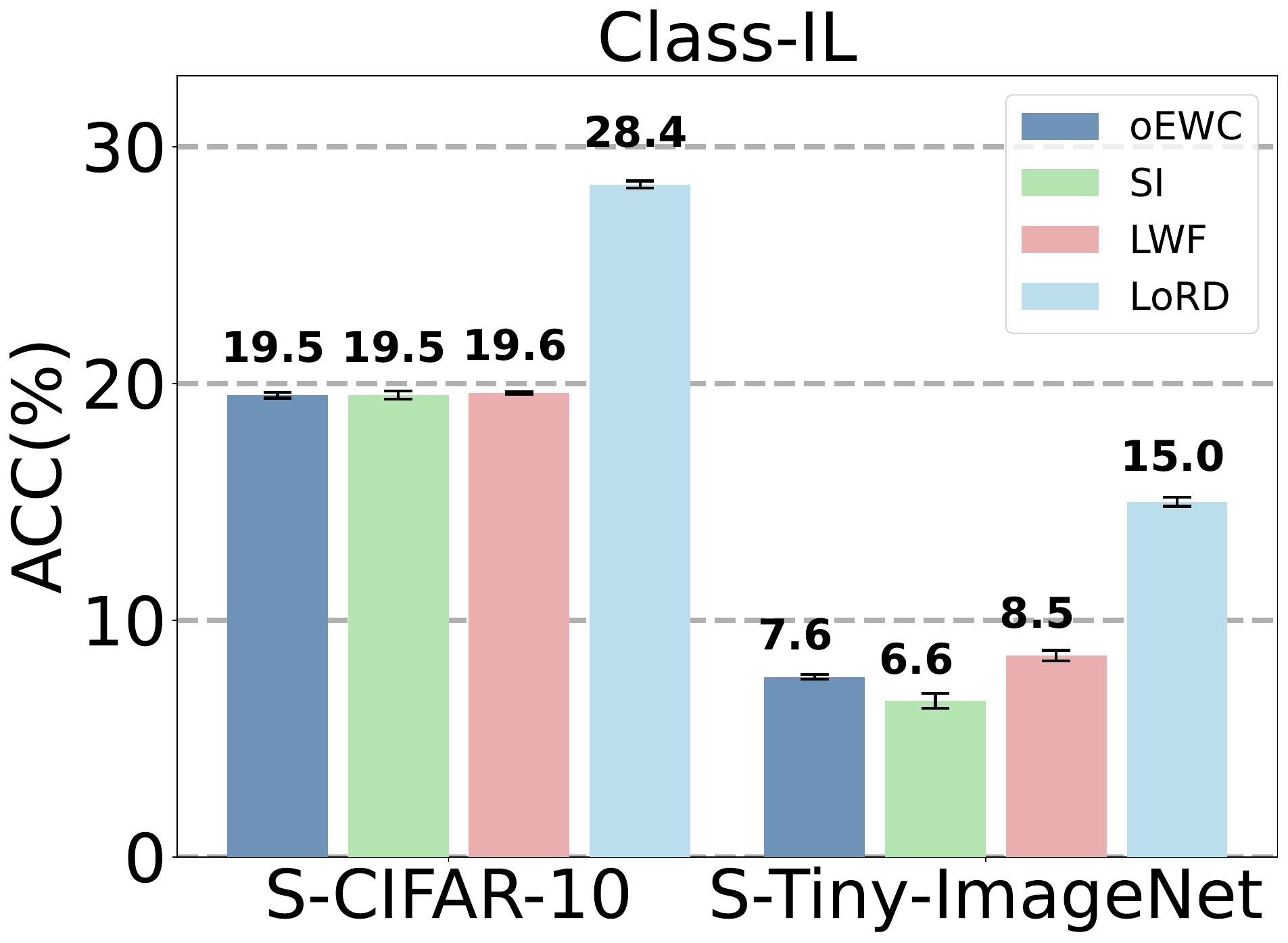}
      \label{fig:Class-IL}
    \end{subfigure}
    \begin{subfigure}[b]{0.4\textwidth}
      \includegraphics[width=\textwidth]{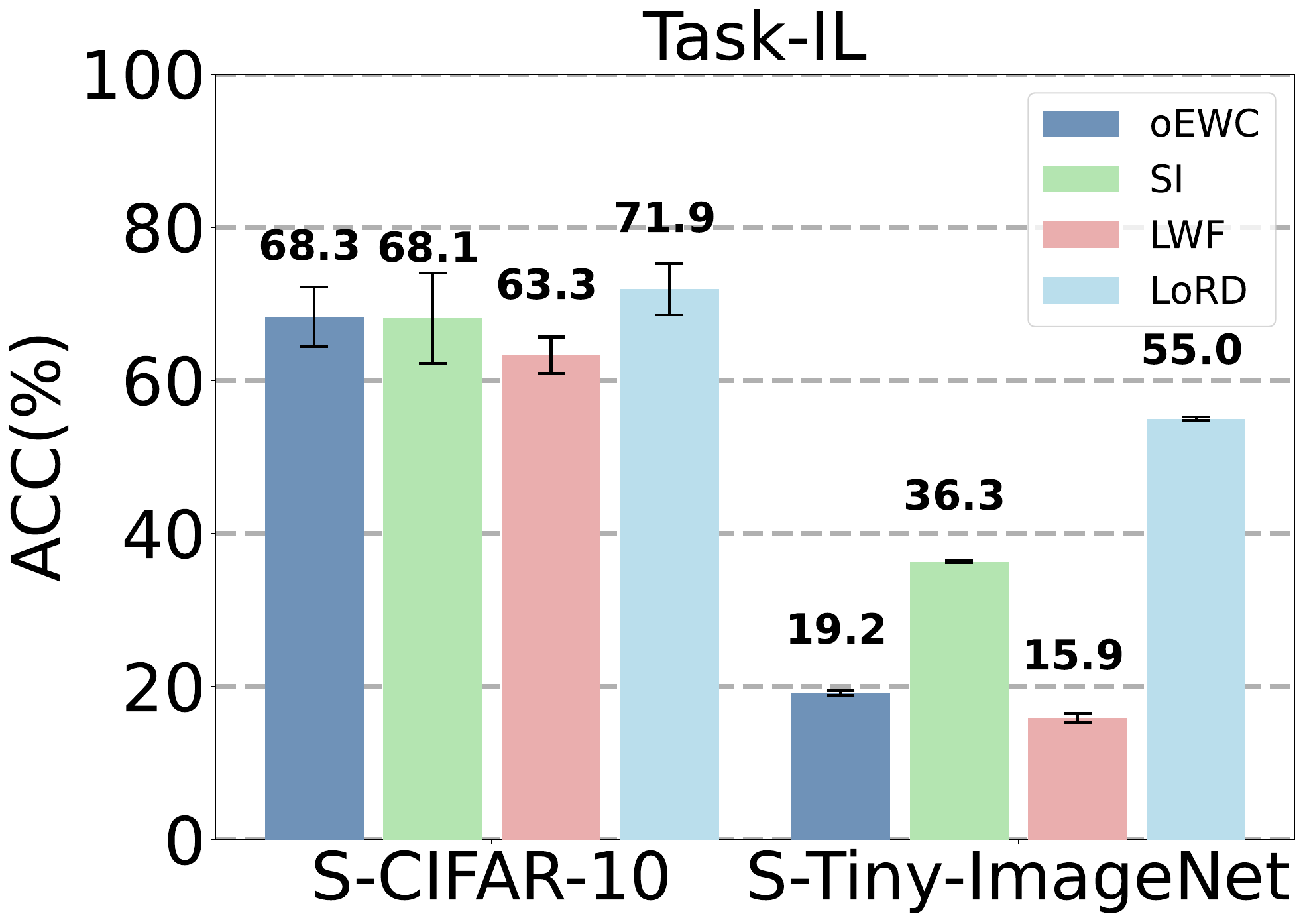}
      \label{fig:Task-IL}
    \end{subfigure}
    \caption{Classification results for standard buffer-free CL benchmarks.}
    \label{fig:combined}
\end{figure}
The results, as illustrated in Fig.~\ref{fig:combined}, present the ACC across all tasks following training.
Traditional regularization-based methods calculate parameter importance based on prior tasks and maintain these critical parameters unaltered. However, as new tasks arrive, the significance of these parameters evolves. In contrast, \method adapts and updates critical parameters during the learning process, continuously optimizing performance and achieving superior results across various scenarios.
In Class-IL, our method surpasses the suboptimal method in ACC on S-CIFAR-10 and S-Tiny-ImageNet datasets by 8.8\% and 6.5\%, respectively.

\begin{figure} [!t]
     \centering
     \includegraphics[width=0.45\linewidth]{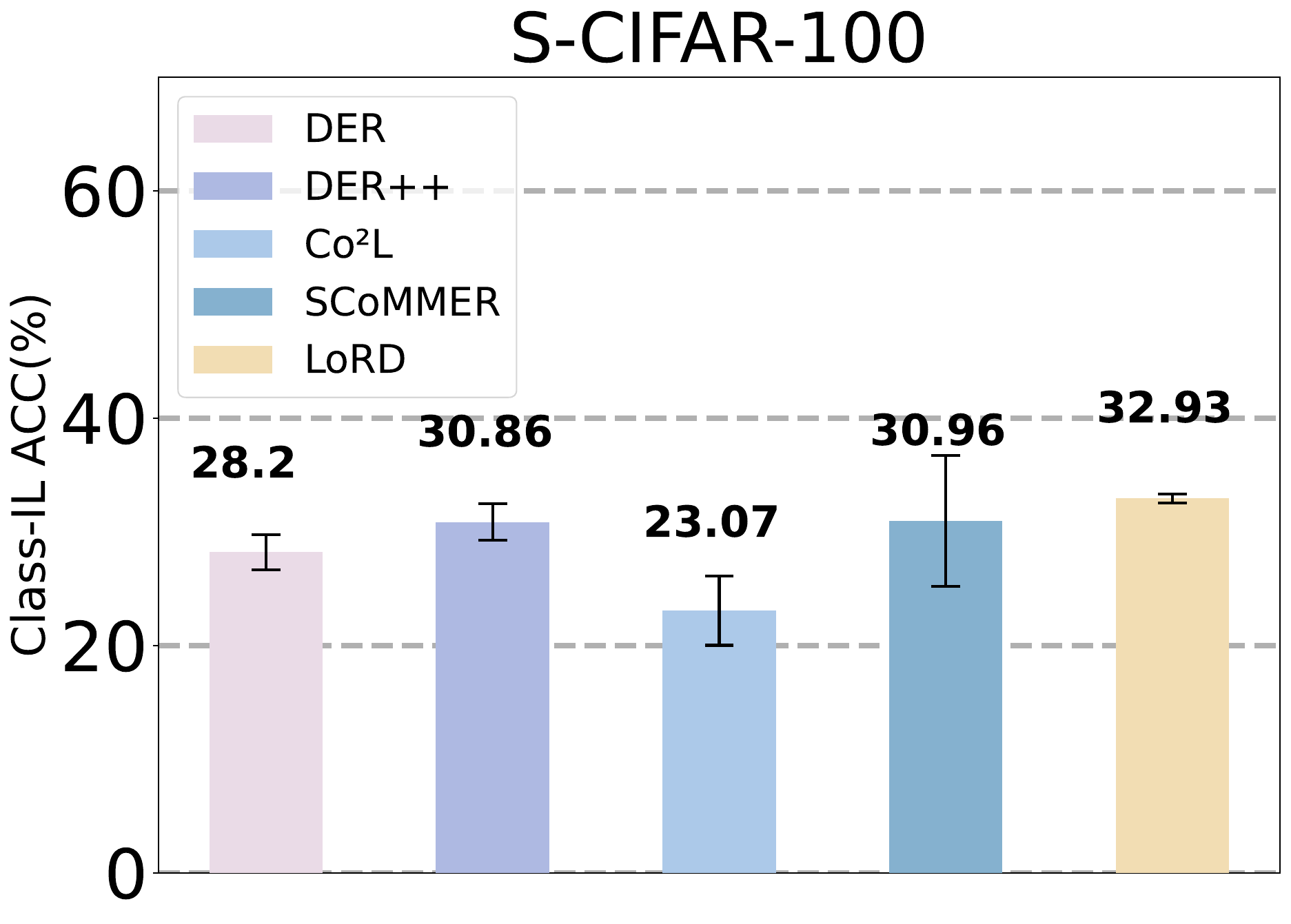}  
     \caption{Results of different methods on S-CIFAR-100 with an unknown number of tasks and a buffer size of 500.}
    \label{fig:n}  
\end{figure}

\paragraph{Settings with Unknown Task Numbers}
To evaluate the performance of \method when the number of tasks $N$ is unknown, we test various methods on the S-CIFAR-100 dataset, which consists of 20 tasks. The results, presented in Fig.~\ref{fig:n}, show that \method achieves the highest Class-IL accuracy, even when $N$ is unknown. This demonstrates that \method effectively mitigates \textit{catastrophic forgetting} and optimizes performance without prior knowledge of the task count.
\paragraph{Forgetting Analysis}
\label{sec:app:Forgetting}
\begin{table}[ht]
	\centering
	\renewcommand\arraystretch{1.2}	\setlength{\tabcolsep}{4mm}
    \small
	\caption{Forgetting results on S-CIFAR-10 dataset.}
	\begin{tabular}{ccccc}
		\hline
		\multirow{2}{*}{{\bf  Method}}&\multicolumn{4}{c}{{\bf S-CIFAR-10}}\\
		\cline{2-5}
		&\multicolumn{2}{c}{Class-IL($\downarrow$)}&\multicolumn{2}{c}{Task-IL($\downarrow$)}\\
		\hline
		\method \textit{w/o} BUF&\multicolumn{2}{c}{28.37}&\multicolumn{2}{c}{71.85}\\
		\hline
		Buffer Size&200&500& 200&500 \\
		\hline
		GEM & 82.61\scriptsize{$\pm$1.60} & 74.31\scriptsize{$\pm$4.62} & 9.27\scriptsize{$\pm$2.07} &  9.12\scriptsize{$\pm$0.21} \\
		\hline
		iCaRL & 28.72\scriptsize{$\pm$0.49}& 25.71\scriptsize{$\pm$1.10}& {\bf 2.63\scriptsize{$\pm$3.48} }& 2.66\scriptsize{$\pm$2.47} \\
		\hline
		ER & 61.24\scriptsize{$\pm$2.62}& 74.35\scriptsize{$\pm$0.07} & 7.08\scriptsize{$\pm$0.64}& 3.54\scriptsize{$\pm$0.35} \\
		\hline
		A-GEM & 95.73\scriptsize{$\pm$0.20} & 94.01\scriptsize{$\pm$1.16} & 16.39\scriptsize{$\pm$0.86} & 14.26\scriptsize{$\pm$4.18} \\
		\hline
		GSS & 75.25\scriptsize{$\pm$4.07} & 62.88\scriptsize{$\pm$2.67} & 8.56\scriptsize{$\pm$1.78} & 7.73\scriptsize{$\pm$3.99} \\
		\hline
		DER & 40.76\scriptsize{$\pm$0.42} & 26.74\scriptsize{$\pm$0.15}& 6.57\scriptsize{$\pm$0.20} & 4.56\scriptsize{$\pm$0.45} \\
		\hline
		DER++ & 32.59\scriptsize{$\pm$4.41}& 22.38\scriptsize{$\pm$0.84} & 5.16\scriptsize{$\pm$1.15} & 4.66\scriptsize{$\pm$0.19} \\
		\hline
		HAL & 69.11\scriptsize{$\pm$4.34} & 62.21\scriptsize{$\pm$7.53} & 12.26\scriptsize{$\pm$1.10} & 5.41\scriptsize{$\pm$0.50} \\
		\hline
		ER-ACE & 35.79\scriptsize{$\pm$5.77} & 24.51\scriptsize{$\pm$1.80} & 6.92\scriptsize{$\pm$0.66} & 4.07\scriptsize{$\pm$0.28} \\
		\hline
            SCoMMER & 23.73\scriptsize{$\pm$4.76} & 16.85\scriptsize{$\pm$1.35} & 6.92\scriptsize{$\pm$1.16} & 2.86\scriptsize{$\pm$0.47} \\
		\hline
		{\bf \method(ours)} & {\bf 23.54\scriptsize{$\pm$3.22}} & {\bf  16.68\scriptsize{$\pm$2.15}} & {\bf 4.19\scriptsize{$\pm$0.20}}  & {\bf  1.84\scriptsize{$\pm$0.31}}   \\
		\hline
	\end{tabular}
	\label{tab:forgetting}
\end{table}
Tab.~\ref{tab:forgetting} presents the FF result. Our method demonstrates the lowest level of forgetting, highlighting its effectiveness in preserving previously acquired knowledge. This indicates that \method not only improves the performance of subsequent tasks but also mitigates the challenge of \textit{catastrophic forgetting}, emphasizing the importance of decoupling task-specific knowledge to maintain stability across tasks.

\begin{table*}[t!]
	\centering
	\renewcommand\arraystretch{1.2}	\setlength{\tabcolsep}{1.1mm}
    \small
	\caption{Comparison of different methods for S-CFAR-100 dataset with a buffer size of 500. $\mathcal{T}$ denotes the number of tasks.}
	\begin{tabular}{ccccccc}
		\hline
		\multirow{2}{*}{{\bf  Method}}&\multicolumn{6}{c}{{\bf S-CIFAR-100}}\\
		\cline{2-7}
		&\multicolumn{3}{c}{Class-IL($\uparrow$)}&\multicolumn{3}{c}{Task-IL($\uparrow$)}\\
		\hline
		Number of Tasks&$\mathcal{T}=20$&$\mathcal{T}=10$&$\mathcal{T}=5$&$\mathcal{T}=20$&$\mathcal{T}=10$&$\mathcal{T}=5$ \\
		\hline
		DER & 28.20\scriptsize{$\pm$1.53}& 33.25\scriptsize{$\pm$2.72} & 40.17\scriptsize{$\pm$2.26}& 78.19\scriptsize{$\pm$0.16} & 74.62\scriptsize{$\pm$0.65} & 70.25\scriptsize{$\pm$0.07} \\
		\hline
		DER++ & 30.86\scriptsize{$\pm$1.62} & 36.68\scriptsize{$\pm$1.25} & 42.34\scriptsize{$\pm$2.82} & 79.54\scriptsize{$\pm$0.78} & 75.49\scriptsize{$\pm$0.36} & 70.81\scriptsize{$\pm$0.19} \\
		\hline
		$Co^2L$ & 23.07\scriptsize{$\pm$3.05} & 27.51\scriptsize{$\pm$1.04} & 29.42\scriptsize{$\pm$4.46}& 61.46\scriptsize{$\pm$5.81} & 54.85\scriptsize{$\pm$3.37} & 49.09\scriptsize{$\pm$6.19} \\
		\hline
		SCoMMER& 30.96\scriptsize{$\pm$5.77} & 38.55\scriptsize{$\pm$5.77} & 47.27\scriptsize{$\pm$5.77} & 84.12\scriptsize{$\pm$5.77}& 80.08 \scriptsize{$\pm$5.77}& 74.75\scriptsize{$\pm$5.77}\\
		\hline
		{\bf \method(Ours)} & {\bf 32.93\scriptsize{$\pm$0.39}} &   {\bf40.61\scriptsize{$\pm$1.24}} & {\bf 48.11\scriptsize{$\pm$0.80}}  &   80.15\scriptsize{$\pm$0.30}  &   76.44\scriptsize{$\pm$0.43} &  73.18\scriptsize{$\pm$0.41}  \\
		\hline
	\end{tabular}
	\label{tab:cifar100}
\end{table*}
\paragraph{Results with Different Task Settings}
\label{sec:app:cifar100}
To evaluate the capacity of various methods for knowledge assimilation, we compared our method against other methods using the S-CIFAR-100 dataset, covering different task settings. Specifically, we evaluate performances in settings of 5, 10, and 20 tasks.
Tab.~\ref{tab:cifar100} presents the performance evaluation of our method on the  S-CIFAR-100 dataset. Across all task settings, our method consistently outperforms other existing methods.
In Class-IL, our \method achieves a 2.06\% and 2\% ACC improvement over the sub-optimal method SCoMMER on the S-CIFAR-100 dataset with 10 and 20 tasks, which is quite notable.
This result highlights that \method not only excels at integrating short-term knowledge but also effectively assimilates long-term knowledge, thereby significantly alleviating forgetting.  This strongly corroborates \method’s robustness across multiple task settings and showcases its remarkable generalization ability.
\subsection{Ablation Study}
\begin{table}[t!]
	\centering
        \small
	\renewcommand\arraystretch{1.2}	\setlength{\tabcolsep}{2.5mm}
     \caption{Ablation studies of \method with 200 buffer size on S-CIFAR-10 dataset.}
	\begin{tabular}{ccccc}
		\hline
		\multirow{2}{*}{{\bf  Method}}&\multicolumn{4}{c}{{\bf S-CIFAR-10}}\\
		\cline{2-5}
		&\multirow{2}{*}{Class-IL($\uparrow$)}&\multirow{2}{*}{Task-IL($\uparrow$)}&\multirow{2}{*}{Forgetting($\downarrow$)}&FLOPs Train\\
        &&&&$\times 10^{15}$($\downarrow$)\\
		\hline
		\method \textit{w/o} BUF& 28.37\scriptsize{$\pm$4.63} & 71.85\scriptsize{$\pm$5.80} & 90.78\scriptsize{$\pm$1.64} &  6.28\\
		\hline
        \method \textit{w/o} $\mathcal{L}_\text{PD}$ & 64.37\scriptsize{$\pm$1.37} & 92.32\scriptsize{$\pm$0.62} & 34.76\scriptsize{$\pm$3.42} &  8.71 \\
		\hline
        \method \textit{w/o} $\mathcal{L}_\text{DER}$ & 67.45\scriptsize{$\pm$2.36} & 92.42\scriptsize{$\pm$0.29} & 25.11\scriptsize{$\pm$0.4.41} &  8.03 \\
		\hline
		\method \textit{w/o} Compression & 65.04\scriptsize{$\pm$4.90} & 91.39\scriptsize{$\pm$1.63} & 26.67\scriptsize{$\pm$4.90} &  16.67 \\
		\hline
		\method \textit{w/o} Pruning & 68.16\scriptsize{$\pm$1.44} & 93.39\scriptsize{$\pm$0.29}& 22.61\scriptsize{$\pm$12.21} & 9.90 \\
		\hline
		\method \textit{w/o} TRS & 68.97\scriptsize{$\pm$2.74} & 93.47\scriptsize{$\pm$0.14} & 29.79\scriptsize{$\pm$5.07} & 9.77 \\
		\hline
		{\bf \method} & {\bf 70.16\scriptsize{$\pm$0.36}} & {\bf 93.94\scriptsize{$\pm$0.36}} &  23.54\scriptsize{$\pm$3.22}  & 9.77   \\
		\hline
	\end{tabular}
	\label{tab:ablation}
\end{table}
\label{sec:exp:Ablation}
Tab.~\ref{tab:ablation} shows the ablation studies for each component in \method on S-CIFAR-10 datasets.
To assess the influence of the replay buffer, we conduct an experimental evaluation by removing the replay buffer component from the \method model (\method \textit{w/o} BUF).
\method \textit{w/o} Compression means that the learnable parameters of the student model are not compressed across tasks, while \method \textit{w/o} Pruning means that the teacher model is not pruned.
From the results, we have the following observations:
(1) The results of \method \textit{w/o} $\mathcal{L}_\text{PD}$ and \method \textit{w/o} $\mathcal{L}_\text{DER}$ indicate that the collaborative use of both distillation losses is crucial for the teacher network to efficiently encode useful knowledge. $\mathcal{L}_\text{PD}$ aims to maintain the integrity of past knowledge by continuously updating the teacher network. $\mathcal{L}_\text{DER}$, on the other hand, focuses on preventing forgetting in both the student and teacher networks, thereby enabling more effective encoding of valuable knowledge. The synergistic effect of these two distillation losses can significantly enhance the effectiveness of model distillation.
(2) The results of \method \textit{w/o} Compression and \method \textit{w/o} Pruning indicate that eliminating redundancy during the distillation process is crucial for achieving a balance between model plasticity and stability, and is also essential for boosting training efficiency.
(3) TRS has been proven to be positive in efficiently selecting saved samples, improving storage utilization, and achieving higher ACC under limited buffer sizes.
(4) \method attains superior ACC by comprehensively deploying all components, thus corroborating the efficacy of its components.\footnote{We provide more extensive ablation studies about each component in Appendix~\ref{sec:app:Ablation}.} 

\begin{figure*} [!t]
     \centering
     \includegraphics[width=1\linewidth]{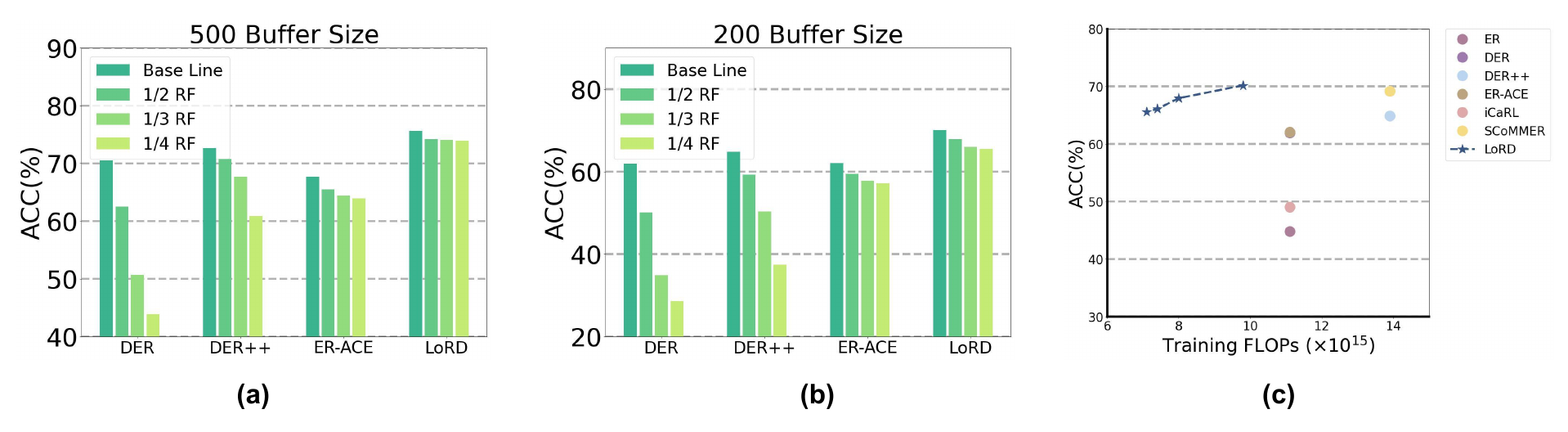}  
     \caption{Results for the quantitative analysis. (a) and (b) illustrate the accuracy at varying rehearsal frequencies. (c) displays the training FLOPs and accuracy of different methods.}
    \label{fig:all}  
\end{figure*}
\begin{table}[t!]
	\centering
        \small
	\renewcommand\arraystretch{1.2}	\setlength{\tabcolsep}{3mm}
     \caption{Comparison of \method and various methods using SparCL on S-CIFAR-10 dataset (Sparsity Ratio: 0.75, Buffer Size: 200).}
	\begin{tabular}{cccc}
		\hline
		\multirow{3}{*}{{\bf  Method}}&\multicolumn{3}{c}{{\bf S-CIFAR-10}}\\
		\cline{2-4}
		&\multirow{2}{*}{Class-IL($\uparrow$)}&\multirow{2}{*}{Task-IL($\uparrow$)}& FLOPs Train \\
		&&&$\times 10^{15}$($\downarrow$)\\
		\hline
		ER-SparCL & 46.89\scriptsize{$\pm$0.68} & 92.02\scriptsize{$\pm$0.72} & 2.0 \\
		\hline
		DER++-SparCL & 66.30\scriptsize{$\pm$0.98} &{\bf 94.06\scriptsize{$\pm$0.45}} & 2.5\\
		\hline
		{\bf\method-SparCL}& {\bf67.07\scriptsize{$\pm$1.65}} & 93.98\scriptsize{$\pm$0.15} &{\bf 1.7}\\
		\hline
	\end{tabular}
	\label{tab:spar}
\end{table}
\subsection{Quantitative Study}
\label{sec:exp:Quantitative}
In this section, we present a comprehensive analysis of our
method \method by conducting a comparative study against existing SOTA CL methods.
\paragraph{Rehearsal Frequency}
\label{sec:exp:RF}
Due to the use of the teacher model, it has become unnecessary to replay in every batch. In our experiment, we decrease the RF to reduce the high computational complexity in rehearsal methods and enhance training efficiency. Fig.~\ref{fig:all} (a) and (b) showcase the ACC of different methods across various RF on the S-CIFAR-10 dataset. When reducing RF, \method exhibits a considerably smaller impact than other methods. At a buffer size of 500 and 200, with the RF adjusted to 1/4, \method demonstrates only a 1.68\% and 4.58\% decrease in ACC, respectively. Additionally, ER-ACE outperforms DER++ and DER due to its cross-entropy computation based solely on the current task’s classes, which enhances model stability and reduces reliance on replay samples. Consequently, \method can dynamically adjust the RF to balance accuracy and efficiency, making it adaptable to various scenarios, including edge computing, with minimal accuracy degradation.
\begin{table}[t!]
	\centering
    \small
	\renewcommand\arraystretch{1.2}	\setlength{\tabcolsep}{3mm}
     \caption{Comparison of training FLOPs across different methods on multiple benchmark datasets.}
	\begin{tabular}{cccc}
		\hline
		{\bf  Method}&{\bf S-CIFAR-10}&{\bf S-CIFAR-100}&{\bf S-Tiny-ImageNet }\\
		\hline
        ER & $11.1\times 10^{15}$ & $11.1\times 10^{15}$ &$17.8\times 10^{16}$ \\
		\hline
        DER++ & $13.9\times 10^{15}$ & $13.9\times 10^{15}$ &$22.2\times 10^{16}$ \\
		\hline
		ER-ACE & $11.1\times 10^{15}$ & $11.1\times 10^{15}$ &$17.8\times 10^{16}$  \\
		\hline
		SCoMMER & $13.9\times 10^{15}$ & $13.9\times 10^{15}$ &$22.2\times 10^{16}$ \\
		\hline
		{\bf \method} & {\bf$9.8\times 10^{15}$} & {\bf$4.6\times 10^{15}$} & {\bf$12.0\times 10^{16}$}  \\
		\hline
	\end{tabular}
	\label{tab:Efficiency}
\end{table}
\paragraph{Training Efficiency}
Minimizing overall training FLOPs is crucial when dealing with data streams to ensure that training keeps pace with the rate at which data becomes available\textcolor{cyan}{~\cite{yu2024ginar}}. Studies\textcolor{cyan}{~\cite{prabhu2023computationally}} have shown that in real-world applications, the ER method performs best due to its high training efficiency. Therefore, when designing CL methods, it is important to ensure that training FLOPs do not exceed those of ER. For comprehensive evaluation, we quantify the cumulative training FLOPs upon completion of terminal tasks across comparative methods. As presented in Tab.~\ref{tab:Efficiency}, our proposed \method consistently achieves the minimal training FLOPs across all benchmark datasets, with particularly pronounced efficiency gains in large-scale task configurations (e.g., 20-task S-CIFAR-100). This enhanced scalability stems from our cosine-annealed model compression mechanism (see Eq.~\eqref{eq:cosine_annealing}), where a higher N value enables the model to maintain elevated compression ratios throughout extended task sequences, thereby achieving FLOPs reduction in prolonged continual learning scenarios. Complementary analysis in Fig.~\ref{fig:all} (c) on S-CIFAR-10 (buffer size=200) reveals that \method achieves the highest ACC with the lowest training FLOPs, occupying the optimal upper-left position in the performance-efficiency landscape. This observation confirms that our compression strategy improves training efficiency without compromising accuracy. Furthermore, by modulating the RF, \method facilitates a flexible balance between computational efficiency and predictive accuracy, rendering it adaptable to diverse application requirements across resource-constrained and precision-sensitive environments.
\begin{table}[t!]
	\centering
    \small
	\renewcommand\arraystretch{1.2}	\setlength{\tabcolsep}{7mm}
     \caption{Comparison of different methods for the S-CIFAR-10 dataset using Nvidia Jetson TX2 with a buffer size of 200 in an online CL setting.}
	\begin{tabular}{ccc}
		\hline
		\multirow{2}{*}{{\bf  Method}}&\multicolumn{2}{c}{{\bf S-CIFAR-10}}\\
		\cline{2-3}
		&\multicolumn{2}{c}{Class-IL($\uparrow$)}\\
		\hline
		Buffer Size&200&500 \\
		\hline
        ER & 37.64\scriptsize{$\pm$15.46} & 45.22\scriptsize{$\pm$9.49} \\
		\hline
        DER++ & 41.93\scriptsize{$\pm$11.94} & 48.04\scriptsize{$\pm$6.39} \\
		\hline
		ER-ACE & 41.18\scriptsize{$\pm$10.79} & 44.30\scriptsize{$\pm$4.79}  \\
		\hline
		SCoMMER & 18.20\scriptsize{$\pm$0.94} & 20.92\scriptsize{$\pm$0.89} \\
		\hline
		{\bf \method} & {\bf 44.92\scriptsize{$\pm$9.30}} & {\bf  48.41\scriptsize{$\pm$5.76}}   \\
		\hline
	\end{tabular}
	\label{tab:tx2}
\end{table}
\paragraph{Sparse Training}
\label{sec:exp:sparse}
To evaluate the compatibility of \method with other efficient CL methods, we train our model using the SOTA sparse training method SparCL~\cite{wang2022sparcl} and test training FLOPs and ACC. We employ a dynamic masking mechanism to selectively mask the weights of the student model. Concurrently, in the weight update phase, a dynamic gradient masking mechanism is utilized to selectively update specific weights during backpropagation. After a certain number of training epochs, data samples that have been well-learned by the model are filtered out, allowing the method to focus more on samples with higher classification difficulty. To address the overfitting issue, we alternate between the distillation method and replay samples in each training batch. Tab.~\ref{tab:spar} demonstrates the results of integrating \method with SparCL. In contrast to the suboptimal DER++ method, \method boosts ACC by 0.77\% and slashes training FLOPs by 32\%. The outstanding performance of \method highlights its effectiveness in eliminating network redundancy. Moreover, the successful integration of \method and SparCL demonstrates \method's adaptability to sparse training, indicating its potential compatibility with pruning methods.

\paragraph{Online CL on Real Edge Devices}
We conduct single-epoch training on real edge devices to simulate real-world online CL scenarios\textcolor{cyan}{~\cite{han2024adaptive}}. Tab.~\ref{tab:tx2} presents the ACC of different methods under the online CL setting on Nvidia Jetson TX2. \method achieves the highest ACC across various buffer sizes. This not only validates its reliability in complex real-world scenarios but also highlights its exceptional scalability and efficiency in resource-constrained edge computing and low-resource environments, indicating its highly promising practical applicability.
\section{Conclusion}
\label{sec:conclusion}
To reduce redundancy in the continual learning process, we draw inspiration from the brain's contextual gating mechanism and propose Low-redundancy Distillation. \method aims to enhance model performance while maintaining training efficiency through collaborative optimization. \method involves pruning the teacher model and compressing the student model's learnable parameters to facilitate the retention and optimization of prior knowledge, effectively decoupling task-specific knowledge without the need to manually assign isolated parameters for each task. Furthermore, we optimize the selection of rehearsal samples and refine RF to improve training efficiency. Extensive experimentation across various benchmark datasets and environments demonstrates \method's superiority, achieving the highest accuracy within the same training time and requiring the minimum training FLOPs to achieve comparable accuracy.

\method is subject to three primary limitations. Firstly, in the context of long task sequences, the student model’s compression strategy applies a higher compression rate to earlier tasks, which, while yielding significant computational benefits, compromises the performance of these initial models. This reflects an inherent trade-off between accuracy and efficiency, necessitating careful consideration of compression parameters to balance these objectives. Secondly, a key limitation of \method is its dependence on predefined task boundaries, an assumption that may not hold in dynamic real-world environments where tasks lack clear delineation. This reliance on prior knowledge of task transitions constrains the method’s adaptability, potentially impairing performance in scenarios with unstable or overlapping task structures. Thirdly, \method is currently tailored exclusively to visual models and specific network architectures, limiting its applicability across diverse domains and model types. Furthermore, \method may exhibit an increased memory footprint when processing large-scale datasets.

To address these limitations, future work can pursue targeted enhancements. For the first limitation, developing adaptive compression strategies that dynamically adjust the compression rate—based on task importance or real-time performance metrics—could preserve early task performance while retaining computational gains. For the second limitation, exploring methods to reduce dependency on predefined task boundaries, such as unsupervised techniques for detecting task shifts, could enhance \method’s flexibility in real-world settings with fluid task structures. For the third limitation, extending \method’s applicability to additional domains, such as natural language processing or reinforcement learning, and adapting it to diverse architectures, including recurrent or transformer-based models, would broaden its utility and generalizability across a wider range of continual learning scenarios.

\appendix
\begin{appendices}
\newtheorem{theorem}{Theorem}
\section{Teacher-aware Reservoir Sampling}
\label{sec:app:TRS}

\subsection{Sample preservation probability in the TRS method}
Here we will use mathematical induction to calculate the probability of retaining the sample in the buffer.
For the first $\mathcal{B}$ samples $x_1 , x_2 , \ldots , x_{\mathcal{B}}$, we will keep them in the buffer, and $p(x_1 \textit{ is retained}) = p(x_2 \textit{ is retained}) = \ldots = p(x_{\mathcal{B}} \textit{ is retained}) = 1$.
For the k-th$(k>\mathcal{B})$ sample, we keep it with a probability of $p(x_k) = \textit{exp}(-\alpha\frac{|\psi_{t}|}{|\theta_{t}|})\frac{\mathcal{B}}{k}$ (which only means it is retained this time).
So the probability of being retained in the first $\mathcal{B}$ samples $x_r(r \in 1:\mathcal{B})$ can be expressed as follows:
\begin{equation}
    \begin{aligned}
    & p(x_r \textit{ is retained})= \\
    & p(x_r \textit{ was retained in the previous round})\times \\
    & (p(x_k \textit{ is discarded})+p(x_k \textit{ is retained}) \times p(x_r \textit{ not replaced}))
    \end{aligned}
    \nonumber 
\end{equation}

\begin{theorem}
    For any positive integer $k (\mathcal{B} < k )$, the following equation holds:
    \[
        p(x_r) = \prod_{n=\mathcal{B}+1}^{k} \frac{n\textit{exp}(\alpha\frac{|\psi_{t}|}{|\theta_{t}|})-1}{n\textit{exp}(\alpha\frac{|\psi_{t}|}{|\theta_{t}|})}
    \]
\end{theorem}

We will prove this equation using mathematical induction.

\textbf{Base Step:} When $k = \mathcal{B}+1$, the sum on the left-hand side is $1 \times ((1-\textit{exp}(-\alpha\frac{|\psi_{t}|}{|\theta_{t}|})\frac{\mathcal{B}}{\mathcal{B}+1}))+\textit{exp}(-\alpha\frac{|\psi_{t}|}{|\theta_{t}|})\frac{\mathcal{B}}{\mathcal{B}+1} \times \frac{\mathcal{B}-1}{\mathcal{B}})=\frac{(\mathcal{B}+1)\textit{exp}(\alpha\frac{|\psi_{t}|}{|\theta_{t}|})-1}{(\mathcal{B}+1)\textit{exp}(\alpha\frac{|\psi_{t}|}{|\theta_{t}|})}$, and the right-hand side evaluates to $\frac{(\mathcal{B}+1)\textit{exp}(\alpha\frac{|\psi_{t}|}{|\theta_{t}|})-1}{(\mathcal{B}+1)\textit{exp}(\alpha\frac{|\psi_{t}|}{|\theta_{t}|})}$ as well. Therefore, the equation holds.

\textbf{Inductive Hypothesis:} Assume that the equation holds for $k = m$, i.e., $p(x_r) = \prod_{n=\mathcal{B}+1}^{m} \frac{n\textit{exp}(\alpha\frac{|\psi_{t}|}{|\theta_{t}|})-1}{n\textit{exp}(\alpha\frac{|\psi_{t}|}{|\theta_{t}|})}$.

\textbf{Inductive Step:} We need to show that the equation also holds for $k = m+1$. We can expand the sum on the left-hand side as:
\begin{equation}
    \begin{aligned}
& p(x_r) = p(x_r \textit{ was retained in the previous round}) \times \\
& ((1-\textit{exp}(-\alpha\frac{|\psi_{t}|}{|\theta_{t}|})\frac{\mathcal{B}}{m+1}))+\\
&\textit{exp}(-\alpha\frac{|\psi_{t}|}{|\theta_{t}|})\frac{\mathcal{B}}{m+1} \times \frac{\mathcal{B}-1}{\mathcal{B}})
    \end{aligned}
    \nonumber 
\end{equation}
Using the inductive hypothesis, $p(x_r \textit{ was retained in the previous round})$ is equal to $\prod_{n=\mathcal{B}+1}^{m} \frac{n\textit{exp}(\alpha\frac{|\psi_{t}|}{|\theta_{t}|})-1}{n\textit{exp}(\alpha\frac{|\psi_{t}|}{|\theta_{t}|})}$. Thus, we can rewrite the above expression as:
\[
    \prod_{n=\mathcal{B}+1}^{m+1} \frac{n\textit{exp}(\alpha\frac{|\psi_{t}|}{|\theta_{t}|})-1}{n\textit{exp}(\alpha\frac{|\psi_{t}|}{|\theta_{t}|})}
\]
Simplifying, we get:
\[
    p(x_r) = \prod_{n=\mathcal{B}+1}^{m+1} \frac{n\textit{exp}(\alpha\frac{|\psi_{t}|}{|\theta_{t}|})-1}{n\textit{exp}(\alpha\frac{|\psi_{t}|}{|\theta_{t}|})}
\]
This proves that the equation holds for $k = m+1$.

By the principle of mathematical induction, the equation holds for any positive integer $n$.

\begin{figure}[t!]
    \vspace{-2mm}
    \centering
    \includegraphics[width=0.65\linewidth]{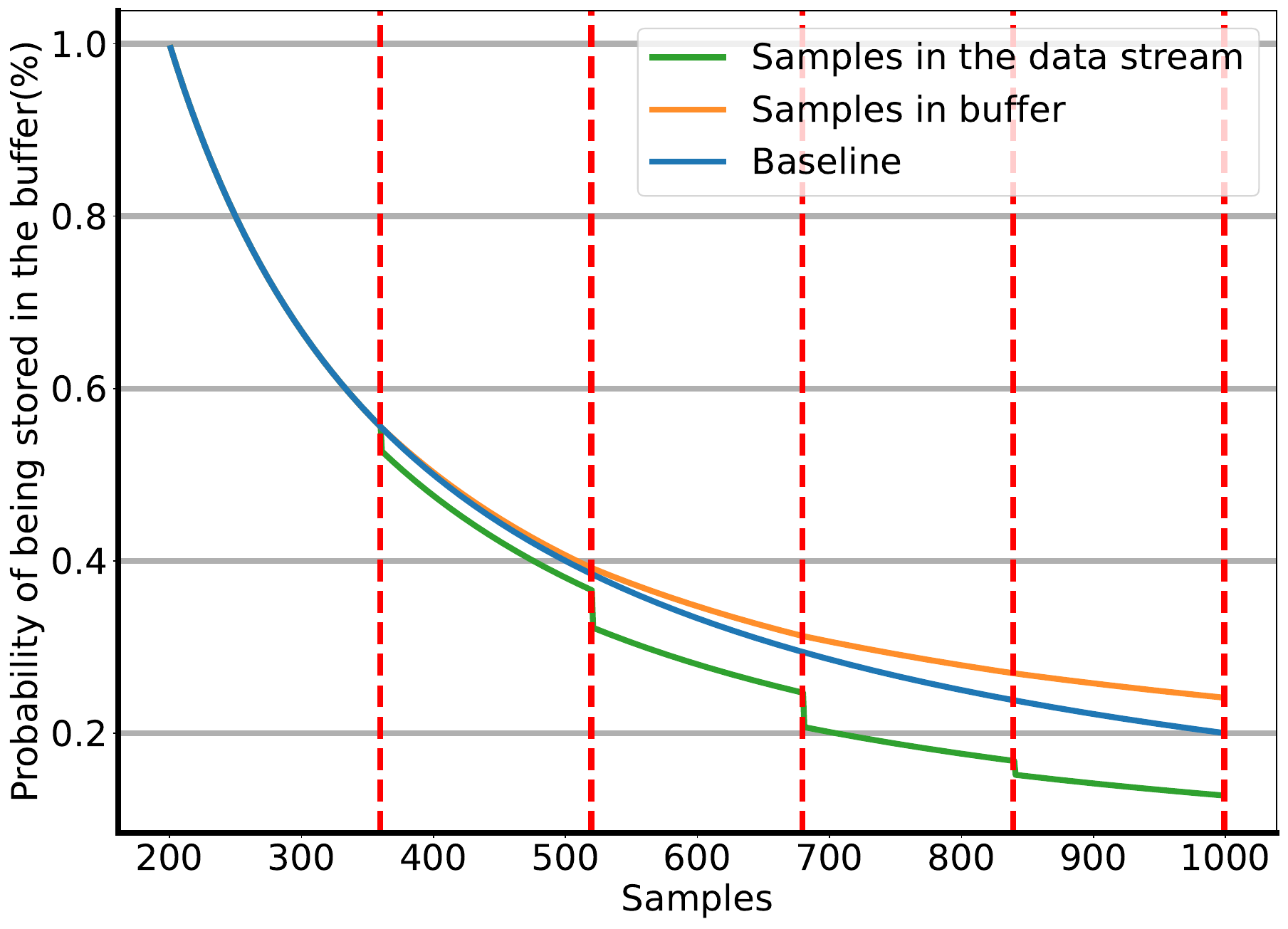}
    \centering
    \caption{
    The probability of different data being stored in the buffer,
    where the \textbf{blue} line represents the reserve sampling method, which ensures an equal probability of storage in the buffer for all data.
    }
    \label{fig:TRS}
    \vspace{-2mm}
\end{figure}
Fig.~\ref{fig:TRS} illustrates the probability of storing individual samples in a buffer under the TRS method, with the parameters $|\mathcal{D}|=1000$ and $\mathcal{B}=200$.

\section{Hyperparameter Selection}
\label{sec:app:Selection}

\begin{table}[h]
\renewcommand\arraystretch{1}	\setlength{\tabcolsep}{0.75mm}
\tiny
\caption{Hyperparameters selected for our experiments.}
\centering\begin{tabular}{| l | c | l | c | l |}
\hline
\textit{Method} & \textit{Buffer} & \textbf{Split Tiny ImageNet} & \textit{Buffer} & \textbf{Split CIFAR-10}  \\
\hline
SGD & - & \textit{lr:} $0.03$  ~~ & - & \textit{lr:} $0.1$ \\
oEWC    & - & \textit{lr:} $0.03$ ~~ \textit{$\lambda$:} $90$ ~~ \textit{$\gamma$:} $1.0$ & - & \textit{lr:} $0.03$ ~~ \textit{$\lambda$:} $10$ ~~ \textit{$\gamma$:} $1.0$ \\
SI      & - & \textit{lr:} $0.03$ ~~ \textit{c:} $1.0$ ~~ \textit{$\xi$:} $0.9$ & - & \textit{lr:} $0.03$ ~~ \textit{c:} $0.5$ ~~ \textit{$\xi$:} $1.0$ \\
LwF     & - & \textit{lr:} $0.01$ ~ \textit{$\alpha$: $1$} ~ \textit{T:} $2.0$ & - & \textit{lr:} $0.03$ ~~ \textit{$\alpha$: $0.5$} ~~ \textit{T:} $2.0$ \\
ER      & 200    & \textit{lr:} $0.1$   & 200  &\textit{lr:} $0.1$  \\
        & 500    & \textit{lr:} $0.03$    & 500  &\textit{lr:} $0.1$  \\
GEM     & & & 200  &\textit{lr:} $0.03$ ~~ \textit{$\gamma$:} $0.5$ \\
        & & & 500  &\textit{lr:} $0.03$ ~~ \textit{$\gamma$:} $0.5$ \\
A-GEM   & 200    & \textit{lr:} $0.01$   & 200  & \textit{lr:} $0.03$  \\
        & 500    & \textit{lr:} $0.01$   & 500  & \textit{lr:} $0.03$  \\
iCaRL   & 200    & \textit{lr:} $0.03$ ~ \textit{wd:} $10^{-5}$ & 200  & \textit{lr:} $0.1$ ~~ \textit{wd:} $10^{-5}$ \\
        & 500    & \textit{lr:} $0.03$ ~~ \textit{wd:} $10^{-5}$ & 500  & \textit{lr:} $0.1$ ~~ \textit{wd:}$10^{-5}$ \\
GSS     & & & 200  & \textit{lr:} $0.03$ ~~ \textit{$gmbs$:} $32$ ~~ \textit{$nb$:} $1$\\
        & & & 500  & \textit{lr:} $0.03$  ~~ \textit{$gmbs$:} $32$ ~~ \textit{$nb$:} $1$\\
HAL     & & & 200  & \textit{lr:} $0.03$ ~~            \textit{$\lambda$:} $0.2$ ~~ \textit{$\beta$:} $0.5$ \\
        & & & & \textit{$\gamma$:} $0.1$ \\
        & & & 500  & \textit{lr:} $0.03$  ~~ \textit{$\lambda$:} $0.1$ ~~ \textit{$\beta$:} $0.3$ \\
        & & & & \textit{$\gamma$:} $0.1$ \\
DER     & 200    & \textit{lr:} $0.03$   ~~ \textit{$\alpha$:} $0.1$ & 200  & \textit{lr:} $0.03$  ~~ \textit{$\alpha$:} $0.3$ \\
        & 500    & \textit{lr:} $0.03$   ~~ \textit{$\alpha$:} $0.1$ & 500  & \textit{lr:} $0.03$  ~~ \textit{$\alpha$:} $0.3$ \\
DER++   & 200    & \textit{lr:} $0.03$  ~~ \textit{$\alpha$:} $0.1$ ~~ \textit{$\beta$:} $1.0$ & 200 & \textit{lr:} $0.03$  ~~ \textit{$\alpha$:} $0.1$ ~~ \textit{$\beta$:} $0.5$ \\
        & 500    & \textit{lr:} $0.03$    ~~ \textit{$\alpha$:} $0.2$ ~~ \textit{$\beta$:} $0.5$ & 500 & \textit{lr:} $0.03$  ~~ \textit{$\alpha$:} $0.2$ ~~ \textit{$\beta$:} $0.5$ \\
C$o^2$L & 200    & \textit{lr:} $0.03$    ~~ \textit{$\kappa$:} $0.1$ ~~ \textit{$\kappa^*$:} $0.1$& 200 & \textit{lr:} $0.03$    ~~ \textit{$\kappa$:} $0.2$ ~~ \textit{$\kappa^*$:} $0.01$ \\
        & &  \textit{$\eta$:} $0.1$ ~~ \textit{$\tau$:} $0.5$ & &\textit{$\eta$:} $0.5$ ~~ \textit{$\tau$:} $0.5$  \\
        & 500    & \textit{lr:} $0.03$    ~~ \textit{$\kappa$:} $0.1$ ~~ \textit{$\kappa^*$:} $0.1$& 500 & \textit{lr:} $0.03$    ~~ \textit{$\kappa$:} $0.2$ ~~ \textit{$\kappa^*$:} $0.01$ \\
        & &  \textit{$\eta$:} $0.1$ ~~ \textit{$\tau$:} $0.5$ & &\textit{$\eta$:} $0.5$ ~~ \textit{$\tau$:} $0.5$ \\
ER-ACE  & 200 & \textit{lr:} $0.1$  ~~  & 200 & \textit{lr:} $0.1$ ~~\\
        & 500 & \textit{lr:} $0.1$ ~~ & 500 & \textit{lr:} $0.1$ ~~\\     
SCoMMER  & 200 & \textit{lr:} $0.1$ ~~ \textit{$\pi_h$:} $0.5$ ~~ \textit{$\pi_s$:} $2$ & 200 & \textit{lr:} $0.1$ ~~ \textit{$\pi_h$:} $0.5$ ~~\textit{$\pi_s$:} $2$\\
& &  \textit{$r$:} $0.1$ & &\textit{$r$:} $0.5$  \\
        & 500 & \textit{lr:} $0.1$ ~~ \textit{$\pi_h$:} $0.5$ ~~ \textit{$\pi_s$:} $2$ & 500 & \textit{lr:} $0.1$ ~~ \textit{$\pi_h$:} $0.5$~~\textit{$\pi_s$:} $2$\\
        & &  \textit{$r$:} $0.15$ & &\textit{$r$:} $0.7$ \\
\method  & 200 & \textit{lr:} $0.1$ ~~ \textit{$\beta_1$:} $1.0$ ~~ \textit{$\beta_2$:} $0.1$ & 200 & \textit{lr:} $0.1$ ~~ \textit{$\beta_1$:} $0.5$ ~~ \textit{$\beta_2$:} $0.1$\\
& &  \textit{$G$:} $19$ ~~ \textit{$\alpha$:} $0.75$ ~~\textit{$\lambda$:} $0.05$& &\textit{$G$:} $10$ ~~ \textit{$\alpha$:} $0.75$ ~~\textit{$\lambda$:} $0.05$  \\
        & 500 & \textit{lr:} $0.1$ ~~ \textit{$\beta_1$:} $0.5$ ~~ \textit{$\beta_2$:} $0.2$ & 500 & \textit{lr:} $0.1$ ~~ \textit{$\beta_1$:} $0.5$ ~~ \textit{$\beta_2$:} $0.2$\\
        & &  \textit{$G$:} $19$ ~~ \textit{$\alpha$:} $0.75$ ~~\textit{$\lambda$:} $0.05$ & &\textit{$G$:} $10$ ~~ \textit{$\alpha$:} $0.75$ ~~\textit{$\lambda$:} $0.05$ \\
        & & & & \\

\hline
\end{tabular}
\centering\begin{tabular}{| l | c | l |}
\hline
\textit{Method} &\textit{Buffer} & \textbf{Split CIFAR-100}  \\
\hline
DER     & 500    & \textit{lr:} $0.03$   ~~ \textit{$\alpha$:} $0.1$ \\
DER++   & 500    & \textit{lr:} $0.03$    ~~ \textit{$\alpha$:} $0.2$ ~~ \textit{$\beta$:} $0.5$ \\
C$o^2$L & 500    & \textit{lr:} $0.03$    ~~ \textit{$\kappa$:} $0.2$ ~~ \textit{$\kappa^*$:} $0.01$ \\
        & &  \textit{$\eta$:} $0.5$ ~~ \textit{$\tau$:} $0.5$  \\
SCoMMER  & 500 & \textit{lr:} $0.1$ ~~ \textit{$\pi_h$:} $0.5$ ~~ \textit{$\pi_s$:} $2$ \\     
& &  \textit{$r$:} $0.1$ \\
\method  & 500 & \textit{lr:} $0.1$ ~~ \textit{$\beta_1$:} $1.0$ ~~ \textit{$\beta_2$:} $0.1$ \\     
& &  \textit{$G$:} $19$ ~~ \textit{$\alpha$:} $0.75$ ~~\textit{$\lambda$:} $0.05$ \\
\hline
\end{tabular}
\label{tab:hyperparams}
\end{table}
Tab.~\ref{tab:hyperparams} presents the selected optimal hyperparameter combinations for each method in the main paper. The hyperparameters include the learning rate (lr), batch size (bs), and minibatch size (mbs) for rehearsal-based methods. Other symbols correspond to specific methods. It should be noted that the batch size and minibatch size are held constant at 32 for all CL benchmarks.

\section{Hyperparameter Sensitivity}
\label{sec:app:Hyperparameter}
\begin{table}[t!]
	\centering
	\renewcommand\arraystretch{0.8}	\setlength{\tabcolsep}{4mm}
	\caption{Average Accuracy of \method on different values of hyperparameters. For a given hyperparameter in the table, all the other hyperparameters are set to their optimal values.}
	\begin{tabular}{ccccc}
		\hline
		\multirow{2}{*}{{\bf  Method}}&\multicolumn{4}{c}{{\bf S-CIFAR-10}}\\
		\cline{2-5}
		&\multicolumn{2}{c}{Class-IL($\uparrow$)}&\multicolumn{2}{c}{Task-IL($\uparrow$)}\\
		\hline
		Buffer Size&200&500& 200&500 \\
		\hline
		$\alpha=0.125$ & 68.16 & 74.92 & 93.41 &  94.74 \\
		\hline
		$\alpha=0.25$ & 68.53 & 74.46 & 93.33 &  94.41 \\
		\hline
		$\alpha=0.375$ & 69.44 & 75.04 & 93.64 &  94.54 \\
		\hline
		$\alpha=0.5$ & 69.33 & 75.68 & 93.65 &  94.81 \\
		\hline
		$\alpha=0.625$ & 68.97 & 75.66 & 93.66 &  94.97 \\
		\hline
		$\alpha=0.75$ & 70.04 & 75.10 & 93.94 &  94.90 \\
		\hline
		$\alpha=0.875$ & 70.06 & 74.95 & 93.71 &  94.93 \\
		\hline
		$\alpha=1$ & 69.22 & 74.90 & 93.84 &  94.79 \\
		\hline
		$\lambda=0.01$ & 68.18 & 75.16 & 93.60 &  94.59 \\
		\hline
		$\lambda=0.025$ & 69.35 & 74.97 & 93.71 &  94.78 \\
		\hline
		$\lambda=0.05$ & 70.16 & 75.66 & 93.75 &  94.95 \\
		\hline
		$\lambda=0.075$ & 68.93 & 75.46 & 93.60 &  94.88 \\
		\hline
		$\lambda=0.2$ & 68.85 & 74.93 & 93.71 &  94.83 \\
		\hline
		$\lambda=0.5$ & 68.90 & 74.85 & 93.88 &  94.39 \\
		\hline
	\end{tabular}
	\label{tab:quantitative}
\end{table}
In Tab.~\ref{tab:quantitative}, we report the performance of \method against a range of hyperparameters.
$\alpha$ (Eq.~\ref{eq:pk} in the main paper) controls the subsequent impact of teacher subnet size on TRS. A larger value of $\alpha$ results in fewer samples being retained for new tasks. $\lambda$ (Eq.~\ref{eq:loss} in the main paper) controls the distillation effect of the teacher subnet on model training. A larger $\lambda$ value reduces the likelihood of the student model forgetting previous tasks, but it also makes learning new tasks more challenging. Experimental results indicate that optimal performance of \method is achieved when parameters $\alpha$ and $\lambda$ are set to 0.75 and 0.05, respectively. More importantly, \method exhibits low sensitivity to hyperparameter selection, significantly enhancing its user-friendliness.

\begin{table}[t!]
	\centering
	\renewcommand\arraystretch{1.2}	\setlength{\tabcolsep}{2mm}
	\caption{Comparison of Different Expansion Methods on S-CIFAR-10 dataset.}
	\begin{tabular}{ccc}
		\hline
		\multirow{2}{*}{{\bf  Method}}&\multicolumn{2}{c}{{\bf S-CIFAR-10}}\\
		\cline{2-3}
		&Class-IL($\uparrow$)&Task-IL($\uparrow$)\\
		\hline
		\method \textit{w/o} Compression & 20.16 & 71.15  \\
		\hline
		\method \textit{w/} Linear & 25.62 & 68.08 \\
		\hline
		{\bf \method \textit{w/} Cos }&{\bf 28.37 }& {\bf71.85} \\
		\hline
	\end{tabular}
	\label{tab:ablation2}
\end{table}
\section{Ablation Study}
\label{sec:app:Ablation}
\begin{figure}[!t]
  \vspace{-2mm}
  \centering
  \includegraphics[width=1\linewidth]{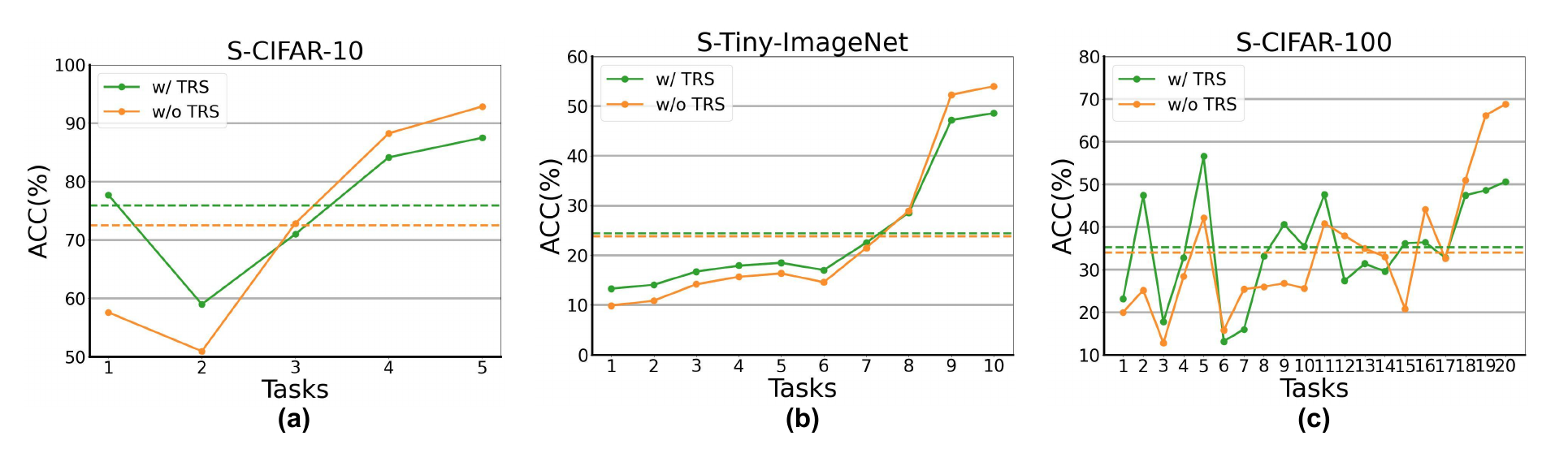}
  \centering
  \caption{ 
  Classification performance across datasets in ablation experiments.
  }
  \label{fig:SCER2}
  \vspace{-2mm}
\end{figure}

\subsection{Student Model Compression}
We study the impact of different parameter allocation methods on model performance without buffer. Tab.~\ref{tab:ablation2} presents the results obtained by allowing all parameters to update (\method \textit{w/o} Compression), linearly allocating parameters (\method \textit{w/} Linear), and cosine allocating parameters (\method \textit{w/} Cos). The results indicate that utilizing cosine parameter allocation greatly enhances the \method's performance. Not restricting parameter updates from the beginning can cause the student model to quickly reach saturation, while linearly allocating parameters may constrain the student model's initial performance. Both scenarios can lead to a decline in the model's overall performance. These experiments provide valuable insights into the significance of student model compression.
\subsection{Teacher model Pruning}
When the teacher model is not pruned, we directly use the student model from the last task as the teacher model for the current task. However, this method does not consider the varying additional parameters required by different tasks. Similar tasks often require fewer additional parameters, indicating a higher degree of similarity between the teacher subnet. On the other hand, tasks with significant differences require more additional parameters, representing larger teacher subnet differences.
By pruning the teacher model, \method can automatically identify the similarity between tasks to improve performance.
\subsection{Teacher-aware Reservoir Sampling}
As depicted in Fig.~\ref{fig:SCER2}, the experimental results showcase the task-specific classification accuracy of the proposed method across various datasets.  The integration of the TRS module enables \method to attain a more balanced and representative learning of prior knowledge, thereby resulting in a substantial enhancement in performance.
\subsection{Different loss function selection}
\label{sec:app:loss}
Our method uses two distillation losses $\mathcal{L}_\text{PD}$ and $\mathcal{L}_\text{DER++}$, where $\mathcal{L}_\text{CE}$ is on all tasks' logits. We conducte additional experiments to verify the efficacy of \method in decoupling task-specific knowledge. First, we calculate $\mathcal{L}_\text{CE}$ solely on the classes of the current task, denoted as $\mathcal{L}_\text{ACE}$. Then, we decompose $\mathcal{L}_\text{DER++}$ into two components: one that only calculates cross-entropy on samples from previous tasks, denoted as $\mathcal{L}_\text{ER}$, and another that aligns the logits of samples from previous tasks, denoted as $\mathcal{L}_\text{DER}$. As shown in Tab.~\ref{tab:loss}, our method yields the highest accuracy by low-redundancy distillation, which restricts the learning capabilities of the distilled subnets for new tasks. Consequently, only the non-distilled subnets can rapidly adapt to new tasks, effectively addressing the task recency bias problem. Meanwhile, the poor performance observed in $\mathcal{L}_\text{ACE}$ further substantiates our viewpoint. This is because \method effectively decouples task-specific knowledge, while calculating the cross-entropy only on the current task's classes would severely impair model's learning on the present task.
\begin{table}[ht]
\centering
\caption{Effect of different loss functions with a
buffer size of 200.}
\begin{tabular}{lcccc}
\hline
\textbf{Loss Function}       & $\mathcal{L}_\text{ACE}$ & $\mathcal{L}_\text{ER}$ & $\mathcal{L}_\text{DER}$ & $\mathcal{L}_\text{DER++}$\\ \hline
\textbf{S-CIFAR-10 Acc} & 55.82              & 67.45                      & 68.25                      & 70.16       \\ \hline
\end{tabular}
\label{tab:loss}
\end{table}

\end{appendices}

\bibliographystyle{unsrt}

\bibliography{refs}



\end{document}